\ifdefined\pdftexversion\pdfoutput=1\relax\fi 
\documentclass[10pt,twocolumn]{article}

\usepackage{arxiv}
\renewenvironment{abstract}{%
  \vspace{6pt}\noindent\textbf{Abstract.}\quad
}{\par\vspace{10pt}}
\usepackage[utf8]{inputenc}
\usepackage[T1]{fontenc}
\usepackage{xurl}
\usepackage{booktabs}
\usepackage{amsmath}
\usepackage{amsfonts}
\usepackage{microtype}
\usepackage{balance}
\usepackage{graphicx}
\usepackage{multirow}
\usepackage{xcolor}
\usepackage{float}
\usepackage{caption}
\usepackage{enumitem}
\usepackage{needspace}
\usepackage[colorlinks=true,linkcolor=blue!60!black,citecolor=blue!60!black,urlcolor=blue!60!black]{hyperref}
\usepackage{cite}
\hypersetup{pdftitle={ChatGPT Images 2.5 in the Wild: A Launch-Period Dataset and Detector Evaluation},
  pdfauthor={Dennis Ng, Xingyu Shen, Ankit Raj, Kidus Zewde, Tommy Duong, Yuchen Zhou, Yuxin Zhang, Neo Tiangratanakul, Simiao Ren}}

\newcommand{\figph}[2]{%
  \IfFileExists{#1}{\includegraphics[width=#2\linewidth]{#1}}%
  {\fbox{\parbox[c][3.4cm][c]{0.95\linewidth}{\centering\textcolor{red}{%
   \texttt{\detokenize{#1}} not generated yet --- see \texttt{figures/README.md}}}}}}

\newcommand{\pmhw}[1]{{\scriptsize\textcolor{black!55}{$\pm$#1}}}

\titleformat{\section}{\large\bfseries\raggedright\hyphenpenalty=10000\exhyphenpenalty=10000}{\thesection}{1em}{}
\titleformat{\subsection}{\normalsize\bfseries\raggedright\hyphenpenalty=10000\exhyphenpenalty=10000}{\thesubsection}{1em}{}
\titlespacing*{\section}{0pt}{12pt plus 2pt minus 2pt}{6pt}
\titlespacing*{\subsection}{0pt}{9pt plus 2pt minus 2pt}{4pt}
\titlespacing*{\paragraph}{0pt}{7pt plus 1pt minus 1pt}{6pt}
\setlist{leftmargin=*,topsep=3pt,itemsep=2pt,parsep=0pt}

\newcommand{\alsoSeenThresh}{4}
\newcommand{\apiCpaDate}{2026-09-09}

\newcommand{\aprilImages}{10,217}
\newcommand{\aprilMedianRegions}{29}
\newcommand{\aprilPctEN}{40.3}
\newcommand{\aprilPctFaces}{59.2}

\newcommand{\aprilPctJA}{32.8}

\newcommand{\aprilPctText}{82.0}

\newcommand{\aprilPctZH}{19.2}
\newcommand{\aprilTweetsRead}{27,662}
\newcommand{\arenaDate}{2026-09-07}
\newcommand{\asOfDate}{2026-09-10}
\newcommand{\auditCost}{3.53}
\newcommand{\auditSeed}{20260909}

\newcommand{\auditorModel}{gpt-5.5}
\newcommand{\balanceAfterTopUp}{100}
\newcommand{\cfApCiHigh}{0.990}
\newcommand{\cfApCiLow}{0.987}

\newcommand{\cfMeasuredAccPct}{89.8}
\newcommand{\cfMeasuredAp}{0.989}
\newcommand{\cfMeasuredMap}{0.947}

\newcommand{\cfPublishedAcc}{0.912}
\newcommand{\cfPublishedMap}{0.994}
\newcommand{\cfShardAp}{0.989}
\newcommand{\cfWorstAuroc}{0.850}
\newcommand{\ciEditFlare}{9}
\newcommand{\ciEditSunburst}{9}
\newcommand{\ciEditTwo}{3}
\newcommand{\ciTtiFlare}{13}
\newcommand{\ciTtiSunburst}{13}
\newcommand{\ciTtiTwo}{4}
\newcommand{\contentN}{3,478}
\newcommand{\costFilter}{4.57}
\newcommand{\costFilterPerK}{0.70}
\newcommand{\costGateway}{0.97}
\newcommand{\costOfficialDayOne}{10.05}

\newcommand{\cpaGeneratorVersion}{2.0}

\newcommand{\eloEditFlare}{1491}
\newcommand{\eloEditSunburst}{1520}
\newcommand{\eloEditTwo}{1461}
\newcommand{\eloTtiFlare}{1399}
\newcommand{\eloTtiSunburst}{1421}
\newcommand{\eloTtiTwo}{1381}
\newcommand{\filterModel}{gpt-5.4-mini}

\newcommand{\filterTokensIn}{4,955,460}
\newcommand{\filterTokensOut}{188,971}
\newcommand{\filterTokensPerImage}{762}
\newcommand{\gatewayCliVersion}{0.1.7}
\newcommand{\gatewayPriceRatio}{67}
\newcommand{\gatewayVendor}{Monid}
\newcommand{\gimBfreeMaxDiff}{1.3}

\newcommand{\gimMaxDelta}{3.5}

\newcommand{\gimNComparisons}{28}
\newcommand{\gimPerGenN}{600}

\newcommand{\gimPgcSaturated}{6}
\newcommand{\gimProbeAdmOurs}{57.0}
\newcommand{\gimProbeAdmPub}{60.0}
\newcommand{\gimProbeBigganOurs}{49.8}
\newcommand{\gimProbeBigganPub}{49.4}
\newcommand{\gimTightDelta}{1.7}
\newcommand{\hashBits}{64}
\newcommand{\hashThresh}{4}
\newcommand{\holdoutDou}{ DoU replicates at +4.7~pp [+2.3,~+7.3] against \xvDouDiff{}~pp.}
\newcommand{\launchStamp}{2026-09-08T18:58:45Z}
\newcommand{\maxFramesPerGif}{11}
\newcommand{\medianTextRegions}{0}
\newcommand{\nAlsoSeen}{259}
\newcommand{\nApiCpaFlare}{371}

\newcommand{\nApiCpaSunburst}{371}
\newcommand{\nApiCpaTotal}{1,007}
\newcommand{\nApiCpaTwo}{265}
\newcommand{\nApiNative}{62}
\newcommand{\nAprilBadge}{4,750}
\newcommand{\nAprilConfirmed}{4,959}
\newcommand{\nAprilImagePosts}{6,606}
\newcommand{\nAprilNameOnly}{508}
\newcommand{\nAprilSample}{3,000}
\newcommand{\nAprilSamplePosts}{2,542}

\newcommand{\nAuditGateAdmPopEN}{377}
\newcommand{\nAuditGateAdmPopJA}{454}
\newcommand{\nAuditGateAdmPopOTHER}{100}
\newcommand{\nAuditGateAdmPopZH}{309}
\newcommand{\nAuditGateAdmitted}{100}

\newcommand{\nAuditGatePerLang}{25}
\newcommand{\nAuditGateRejPopEN}{510}
\newcommand{\nAuditGateRejPopJA}{174}
\newcommand{\nAuditGateRejPopOTHER}{82}
\newcommand{\nAuditGateRejPopZH}{221}

\newcommand{\nAuditLang}{150}
\newcommand{\nAuditLangPer}{50}
\newcommand{\nAuditScreen}{300}
\newcommand{\nAuditScreenChatShot}{65}
\newcommand{\nAuditScreenCollage}{65}

\newcommand{\nAuditScreenPopChatShot}{309}
\newcommand{\nAuditScreenPopCollage}{302}
\newcommand{\nAuditScreenPopOther}{61}
\newcommand{\nAuditScreenPopPromo}{818}
\newcommand{\nAuditScreenPopScreenPhoto}{38}
\newcommand{\nAuditScreenPopStandalone}{1,382}
\newcommand{\nAuditScreenPopulation}{2,910}
\newcommand{\nAuditScreenPromo}{66}
\newcommand{\nAuditScreenScreenPhoto}{38}

\newcommand{\nAuditScreenSrcIg}{9}
\newcommand{\nAuditScreenSrcNote}{6}
\newcommand{\nAuditScreenSrcPx}{4}
\newcommand{\nAuditScreenSrcTw}{217}
\newcommand{\nAuditScreenSrcWb}{5}
\newcommand{\nAuditScreenSrcXhs}{59}
\newcommand{\nAuditScreenStandalone}{66}
\newcommand{\nAuditTax}{100}

\newcommand{\nAuditTaxTw}{90}

\newcommand{\nCallsIg}{7}
\newcommand{\nCallsRd}{116}
\newcommand{\nCallsTwGateway}{273}
\newcommand{\nCallsWb}{324}
\newcommand{\nCallsXhs}{212}
\newcommand{\nCaptionTier}{1,749}

\newcommand{\nCfEvalGenerators}{21}
\newcommand{\nCfEvalImages}{10,000}
\newcommand{\nCfEvalPerLabel}{5,000}
\newcommand{\nCfEvalPopulation}{51,836}
\newcommand{\nChatExport}{85}

\newcommand{\nCollectorKeptRd}{10}

\newcommand{\nCpaFound}{2}
\newcommand{\nCpaProbed}{5}
\newcommand{\nCsvColumns}{22}
\newcommand{\nDayOnePostsInPilot}{139}

\newcommand{\nDayOnePostsLostEntirely}{6}
\newcommand{\nDayOnePostsSurviving}{160}

\newcommand{\nDistinctSizes}{91}
\newcommand{\nDupCrossPost}{711}
\newcommand{\nDupGifFrame}{27}
\newcommand{\nDupImage}{1,281}
\newcommand{\nDupKey}{274}
\newcommand{\nDupSamePost}{570}
\newcommand{\nDupSamePostCrossRun}{302}

\newcommand{\nDupSamePostSameRun}{268}
\newcommand{\nDupSamePostSameRunGif}{13}
\newcommand{\nDupSamePostSameRunStills}{255}

\newcommand{\nDupSamePostSameRunStillsX}{38}

\newcommand{\nFaces}{2,052}
\newcommand{\nFetchedIg}{97}

\newcommand{\nFetchedRd}{193}

\newcommand{\nFilterCalls}{6,490}

\newcommand{\nGateAdmOfficialEN}{11}
\newcommand{\nGateAdmOfficialJA}{8}
\newcommand{\nGateAdmOfficialOTHER}{2}
\newcommand{\nGateAdmOfficialZH}{3}

\newcommand{\nGateRejYesEN}{10}
\newcommand{\nGateRejYesJA}{15}
\newcommand{\nGateRejYesOTHER}{4}
\newcommand{\nGateRejYesZH}{12}
\newcommand{\nGatedLangEN}{1,082}

\newcommand{\nGatedLangJA}{1,265}
\newcommand{\nGatedLangNull}{1,612}
\newcommand{\nGatedLangOther}{127}
\newcommand{\nGatedLangZH}{2,258}
\newcommand{\nGatewayCalls}{932}
\newcommand{\nGatewayEndpoints}{5}
\newcommand{\nGatewayNotObservedOfficial}{1,022}

\newcommand{\nGatewayPosts}{1,524}
\newcommand{\nGatewayReturnedByOfficial}{502}
\newcommand{\nGatewayReturnedFailedGate}{334}
\newcommand{\nGatewayReturnedFailedGateRegex}{318}

\newcommand{\nGenImageGenerators}{7}
\newcommand{\nGenImageImages}{4,200}
\newcommand{\nGenImagePerClass}{300}
\newcommand{\nGifAnimations}{17}

\newcommand{\nGifFrame}{69}
\newcommand{\nGimSelfReported}{4}
\newcommand{\nHoldoutImages}{7,217}
\newcommand{\nHoldoutPosts}{5,153}
\newcommand{\nHoldoutSharedImages}{1,894}
\newcommand{\nHoldoutSharedPosts}{1,089}
\newcommand{\nHostTier}{1,358}
\newcommand{\nHostVideoTier}{12}

\newcommand{\nIgPromo}{140}

\newcommand{\nImages}{3,478}
\newcommand{\nImagineNewest}{500}
\newcommand{\nImagineStyleFlare}{41901}
\newcommand{\nImagineStyleSunburst}{41902}
\newcommand{\nKeptIg}{40}
\newcommand{\nKeptNightcafe}{1,358}
\newcommand{\nKeptNote}{137}

\newcommand{\nKeptPostsPx}{18}

\newcommand{\nKeptPx}{66}
\newcommand{\nKeptRd}{11}
\newcommand{\nKeptTw}{1,501}
\newcommand{\nKeptWb}{40}
\newcommand{\nKeptXhs}{325}
\newcommand{\nLabChatShot}{502}
\newcommand{\nLabCollage}{506}
\newcommand{\nLabOther}{117}
\newcommand{\nLabPromo}{1,661}
\newcommand{\nLabScreenPhoto}{64}
\newcommand{\nLabStandalone}{3,494}
\newcommand{\nLabeled}{6,344}

\newcommand{\nLangCountReal}{16}
\newcommand{\nLangEN}{675}
\newcommand{\nLangFR}{19}
\newcommand{\nLangJA}{703}
\newcommand{\nLangNull}{1,409}

\newcommand{\nLangOtherAll}{60}

\newcommand{\nLangTailReal}{12}
\newcommand{\nLangZH}{631}
\newcommand{\nLateImages}{1,667}
\newcommand{\nLateNightCafe}{1,358}

\newcommand{\nLooseTierNoCreation}{1,296}

\newcommand{\nLooseTierWouldPass}{372}

\newcommand{\nMageFlareTwo}{72}

\newcommand{\nMageSunburstTwo}{149}
\newcommand{\nMageTotal}{117}
\newcommand{\nMageTotalTwo}{221}
\newcommand{\nMerged}{6,356}
\newcommand{\nMergedImages}{6,344}
\newcommand{\nMergedPosts}{3,624}
\newcommand{\nNamedTwoFiveIg}{65}

\newcommand{\nNamedTwoFiveRd}{88}

\newcommand{\nNcCreations}{1,449}

\newcommand{\nNightCafeSampled}{240}

\newcommand{\nNoImageRd}{78}

\newcommand{\nOfficialGatewayOverlap}{168}

\newcommand{\nOfficialReturnedPosts}{1,292}
\newcommand{\nOfficialStrictPosts}{305}
\newcommand{\nOfficialStrictRecords}{594}
\newcommand{\nOfficialStrictTweets}{305}
\newcommand{\nOfficialTier}{371}

\newcommand{\nPhoto}{3,409}
\newcommand{\nPilotStrictPosts}{139}
\newcommand{\nPlatformDefaultJA}{203}
\newcommand{\nPlatformDefaultLang}{568}
\newcommand{\nPlatformDefaultZH}{365}
\newcommand{\nPlatforms}{8}

\newcommand{\nPostLaunchRd}{176}

\newcommand{\nPosts}{2,440}

\newcommand{\nPostsEN}{364}
\newcommand{\nPostsJA}{360}
\newcommand{\nPostsZH}{299}

\newcommand{\nPxAiDeclared}{66}

\newcommand{\nPxCpaFound}{31}
\newcommand{\nPxCpaNamesTwo}{30}

\newcommand{\nPxCpaTotal}{66}

\newcommand{\nRecords}{10,163}

\newcommand{\nReviewedExcluded}{16}

\newcommand{\nRoutesWord}{nine}

\newcommand{\nSeaArtCards}{2}
\newcommand{\nSeaArtTasksFlare}{1,161}
\newcommand{\nSeaArtTasksSunburst}{1,631}

\newcommand{\nSeaArtWorksTotal}{0}
\newcommand{\nSeenIg}{193}

\newcommand{\nSeenRd}{18}

\newcommand{\nSilentTotal}{106}
\newcommand{\nSilentTwo}{35}

\newcommand{\nSrcFive}{2}
\newcommand{\nSrcFour}{4}

\newcommand{\nSrcThree}{36}
\newcommand{\nSrcTwo}{217}
\newcommand{\nStrict}{7,911}
\newcommand{\nStrictTierWidens}{5}

\newcommand{\nTaxAuditorPortrait}{29}
\newcommand{\nTaxAuditorTextGraphic}{6}

\newcommand{\nTaxClipAnime}{44}

\newcommand{\nTaxClipPortrait}{5}
\newcommand{\nTaxClipTextGraphic}{17}

\newcommand{\nText}{1,357}

\newcommand{\nTweetsRead}{2,009}
\newcommand{\nVideo}{12}

\newcommand{\nWithDims}{371}

\newcommand{\nWordEligiblePosts}{1,082}

\newcommand{\nXImages}{1,501}
\newcommand{\nXLang}{1,501}
\newcommand{\nXLangEN}{675}
\newcommand{\nXLangJA}{500}
\newcommand{\nXLangZH}{266}

\newcommand{\nXPosts}{861}

\newcommand{\ncFirstSeen}{2026-09-10}
\newcommand{\ncWindowEnd}{2026-09-10 22:06Z}
\newcommand{\ncWindowHours}{16.7}
\newcommand{\ncWindowStart}{2026-09-10 05:25Z}
\newcommand{\noLangSourceList}{Instagram and Reddit and NightCafe}
\newcommand{\pctAlsoSeen}{7.4}
\newcommand{\pctArtHigh}{96.5}
\newcommand{\pctArtLow}{1.5}

\newcommand{\pctCfGenPrevalence}{5}

\newcommand{\pctClsFantasy}{27.2}

\newcommand{\pctCreationAll}{67.3}
\newcommand{\pctCreationAllHi}{74.3}
\newcommand{\pctCreationAllLo}{59.5}
\newcommand{\pctCreationEN}{66.0}

\newcommand{\pctCreationJA}{82.0}

\newcommand{\pctCreationZH}{54.0}
\newcommand{\pctCreationZHHi}{67.0}
\newcommand{\pctCreationZHLo}{40.4}

\newcommand{\pctDouSeedShift}{0.43}

\newcommand{\pctFaces}{59.0}
\newcommand{\pctFantasyWithoutNc}{10.3}
\newcommand{\pctFprHigh}{5.9}
\newcommand{\pctFprLow}{3.5}
\newcommand{\pctGatePrec}{64.0}

\newcommand{\pctGatePrecHi}{72.7}

\newcommand{\pctGatePrecLo}{54.2}

\newcommand{\pctGatewayInOfficial}{11}
\newcommand{\pctGatewayReturnedByOfficial}{33}

\newcommand{\pctGenFprHigh}{4.0}
\newcommand{\pctGenRecallHigh}{100}
\newcommand{\pctGenRecallLow}{47}

\newcommand{\pctLandscape}{23.7}
\newcommand{\pctLangAgreeAll}{99.3}

\newcommand{\pctLangAgreeEN}{100.0}

\newcommand{\pctLangAgreeJA}{100.0}

\newcommand{\pctLangAgreeZH}{98.0}

\newcommand{\pctLangNull}{40.5}

\newcommand{\pctLateNightCafe}{81}

\newcommand{\pctLooseTierWouldPass}{21.3}
\newcommand{\pctNcFantasy}{53.6}
\newcommand{\pctNcFantasyShare}{77.0}
\newcommand{\pctNcImageShare}{39.0}

\newcommand{\pctNoImageRd}{89}

\newcommand{\pctOfficialInGateway}{55}
\newcommand{\pctPgcArt}{96.5}
\newcommand{\pctPgcPhoto}{5.1}
\newcommand{\pctPlatformDefaultLang}{16.3}
\newcommand{\pctPortrait}{51.5}
\newcommand{\pctProbeWild}{3.7}
\newcommand{\pctPromoXGateway}{21.4}
\newcommand{\pctPromoXOfficial}{11.1}
\newcommand{\pctPxCpa}{47}

\newcommand{\pctRetainAll}{54.8}
\newcommand{\pctRetainLangEN}{62.4}

\newcommand{\pctRetainLangJA}{55.6}
\newcommand{\pctRetainLangNull}{87.4}
\newcommand{\pctRetainLangOther}{47.2}
\newcommand{\pctRetainLangZH}{27.9}

\newcommand{\pctScreenAgreeRaw}{72.3}
\newcommand{\pctScreenContam}{1.5}
\newcommand{\pctScreenContamHi}{8.1}
\newcommand{\pctScreenContamLo}{0.3}

\newcommand{\pctScreenPrecChatShot}{90.8}

\newcommand{\pctScreenPrecCollage}{69.2}
\newcommand{\pctScreenPrecCollageHi}{79.1}
\newcommand{\pctScreenPrecCollageLo}{57.2}
\newcommand{\pctScreenPrecPromo}{62.1}
\newcommand{\pctScreenPrecPromoHi}{72.9}
\newcommand{\pctScreenPrecPromoLo}{50.1}
\newcommand{\pctScreenPrecScreenPhoto}{18.4}

\newcommand{\pctScreenPrecStandalone}{98.5}
\newcommand{\pctScreenPrecStandaloneHi}{99.7}
\newcommand{\pctScreenPrecStandaloneLo}{91.9}

\newcommand{\pctScreenRemoved}{44.9}

\newcommand{\pctSquare}{24.8}
\newcommand{\pctStandaloneAll}{55.1}

\newcommand{\pctTaxAgree}{61.0}

\newcommand{\pctTaxAnime}{68.2}

\newcommand{\pctTaxMockup}{37.5}

\newcommand{\pctTaxTextGraphic}{23.5}

\newcommand{\pctText}{39.0}
\newcommand{\pctTwFantasy}{11.3}
\newcommand{\pctWildFlagHigh}{56.4}
\newcommand{\pctWildFlagLow}{3.7}
\newcommand{\pctWithDims}{10.7}
\newcommand{\pctWordConsistency}{6.8}
\newcommand{\pctWordEdit}{10.5}
\newcommand{\pctWordFlare}{6.7}
\newcommand{\pctWordPrompt}{31.6}
\newcommand{\pctWordSpeed}{6.7}
\newcommand{\pctWordSunburst}{10.4}
\newcommand{\ppGapHigh}{81}
\newcommand{\ppGapLow}{42}
\newcommand{\ppGapOodHigh}{80}
\newcommand{\ppGapOodLow}{22}
\newcommand{\ppGapStrictHigh}{82}
\newcommand{\ppGapStrictLow}{45}
\newcommand{\priceFilterIn}{0.75}
\newcommand{\priceFilterOut}{4.50}
\newcommand{\priceGatewayPage}{0.0015}
\newcommand{\priceOfficialTweet}{0.005}
\newcommand{\priceXhsPage}{0.015}
\newcommand{\probeDateOne}{2026-09-09}
\newcommand{\probeDateTwo}{2026-09-10}

\newcommand{\screenKappa}{0.65}
\newcommand{\secsIg}{11.9}
\newcommand{\secsRd}{4.7}
\newcommand{\secsTwGateway}{6.6}
\newcommand{\secsWb}{3.8}
\newcommand{\secsXhs}{3.9}

\newcommand{\taxKappa}{0.51}

\newcommand{\unsafeFlare}{1.41}
\newcommand{\unsafeSunburst}{1.09}
\newcommand{\unsafeTwo}{1.64}
\newcommand{\votesEditFlare}{5,676}
\newcommand{\votesEditSunburst}{6,704}
\newcommand{\votesEditTwo}{235,928}

\newcommand{\windowEnd}{2026-09-10 22:06:41Z}
\newcommand{\windowHours}{51.1}
\newcommand{\windowStart}{2026-09-08 18:58:45Z}
\newcommand{\xvDouBonfHi}{+7.2}
\newcommand{\xvDouBonfLo}{+0.2}
\newcommand{\xvDouDiff}{+3.6}
\newcommand{\xvDouHi}{+6.2}
\newcommand{\xvDouHoldBonfLo}{+2.0}

\newcommand{\xvDouLo}{+1.0}
\newcommand{\xvDraws}{50,000}
\newcommand{\xvEffortBonfHi}{+10.8}
\newcommand{\xvEffortBonfLo}{+1.6}
\newcommand{\xvEffortDiff}{+6.1}
\newcommand{\xvEffortHi}{+9.6}
\newcommand{\xvEffortHoldBonfLo}{+2.1}
\newcommand{\xvEffortHoldDiff}{+5.8}
\newcommand{\xvEffortHoldHi}{+9.1}
\newcommand{\xvEffortHoldLo}{+2.6}
\newcommand{\xvEffortLo}{+2.7}
\newcommand{\xvMaxHalfWidth}{3.9}
\newcommand{\xvNDetectors}{6}
\newcommand{\xvNSignificant}{2}

\newcommand{\calEffortBonfLo}{+1.7}
\newcommand{\calEffortBonfHi}{+11.3}
\newcommand{\calDouBonfLo}{+0.3}
\newcommand{\calDouBonfHi}{+7.4}

\newcommand{\keepdate}[1]{\expandafter\let\csname raw\string#1\endcsname#1%
  \expandafter\def\expandafter#1\expandafter{\expandafter\mbox\expandafter{\csname raw\string#1\endcsname}}}
\keepdate\apiCpaDate \keepdate\arenaDate \keepdate\asOfDate \keepdate\ncFirstSeen
\keepdate\probeDateOne \keepdate\probeDateTwo \keepdate\ncWindowStart \keepdate\ncWindowEnd
\keepdate\windowStart \keepdate\windowEnd \keepdate\launchStamp
\title{ChatGPT Images 2.5 in the Wild:\\
A Launch-Period Dataset and Detector Evaluation}

\author{%
  Dennis Ng\textsuperscript{*} \quad Xingyu Shen\textsuperscript{*} \quad Ankit Raj\textsuperscript{*} \quad
  Kidus Zewde\textsuperscript{*} \quad Tommy Duong\textsuperscript{*}\\[1pt]
  Yuchen Zhou\textsuperscript{*} \quad Yuxin Zhang\textsuperscript{*} \quad
  Neo Tiangratanakul\textsuperscript{*} \quad Simiao Ren\textsuperscript{\dag}\\[2pt]
  Scam.ai\\[1pt]
  {\footnotesize \textsuperscript{*}Equal contribution. \quad \textsuperscript{\dag}Corresponding author:
  \texttt{benren@scam.ai}}%
}
\date{}

\begin{document}
\maketitle
{\renewcommand{\thefootnote}{}\footnote{Authorship of this paper does not imply employment by or affiliation with Scam.AI (Reality Inc.); some authors are independent researchers who collaborated on this work.}\addtocounter{footnote}{-1}}

\begin{abstract}
An image tool can change its underlying generator while retaining its public
name, making version attribution from online posts ambiguous. We study this
problem after the ChatGPT Images~2.5 launch. Our frozen collection contains
\nImages{} images from \nPosts{} posts across \nPlatforms{} sources.
Recorded posting times fall within the first \windowHours{} hours after the
announcement. It records three
attribution tiers and retains standalone images after image-form filtering
and targeted review. Caption claims and host records provide admission
evidence, not independently verified generator identity.

The observed content profile depends on the source mixture: NightCafe supplies
\pctNcImageShare\% of images but \pctNcFantasyShare\% of CLIP-assigned fantasy
scenes. We then evaluate six frozen detectors at thresholds calibrated to a
5\% flag rate on reference photographs. Collection flag rates range from
\pctWildFlagLow{} to \pctWildFlagHigh\%, falling
\ppGapLow{}--\ppGapHigh{} percentage points below GenImage recall.
Held-out artwork false-positive rates range from \pctArtLow{} to
\pctArtHigh\%, so a higher collection flag rate does not by itself establish
better detection.

An exploratory X-only comparison with our April collection finds a higher
September flag rate for Effort, and a suggestive difference for DoU, under
fixed-threshold post-clustered bootstrap intervals. Attribution, content and
processing differences prevent a causal interpretation of these contrasts. The
collection supports analysis of reported model use during a product
transition, with source and attribution evidence retained for interpretation.
The collection is released at \mbox{\url{https://scam.ai/research}}.
\end{abstract}


\begin{figure*}[t]
  \centering
  \includegraphics[width=\textwidth]{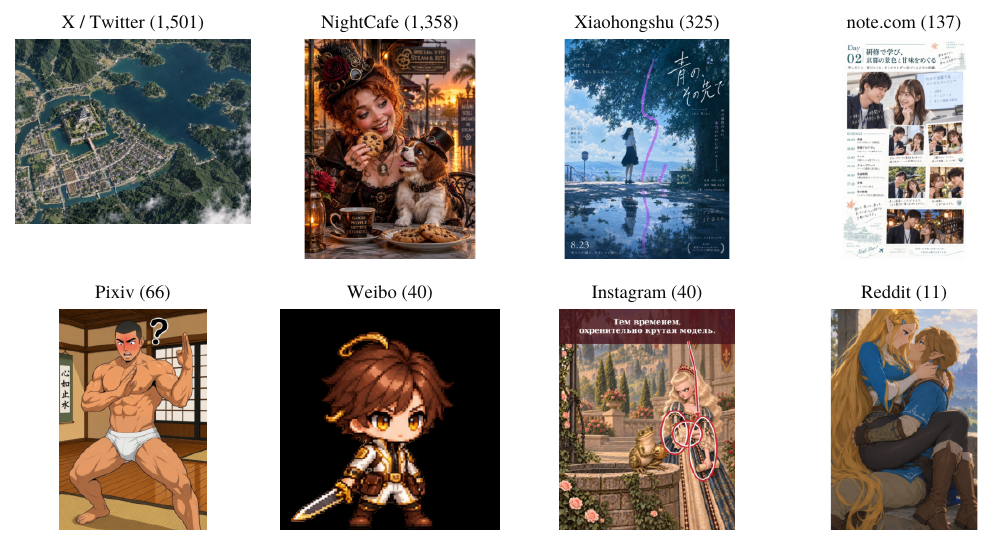}
  \caption{One illustrative image from each of the eight sources in the frozen
  collection. Labels give each source's full image count. Examples are selected
  from the previously fixed illustration set and preserve their aspect ratios.
  This source-balanced display is not a random or proportionally representative
  sample; Figure~\ref{fig:platform_content} shows subject distributions within
  each source.}
  \label{fig:sample_grid}
\end{figure*}

\section{Introduction}
\label{sec:intro}

A model-specific dataset collected from online posts depends on evidence
that connects each image to the claimed generator. That link becomes
ambiguous when a product changes its underlying model while retaining its
public name. A caption naming the product may then refer to either version,
and a post made after an upgrade can still contain an older image.

OpenAI released ChatGPT Images~2.5 on 8 September 2026 through
\texttt{gpt-image-2.5-flare} and \texttt{gpt-image-2.5-sunburst}, alongside
an upgrade to ChatGPT's image tool~\cite{openai2026images25}.
Our April GPT-Image-2 collection~\cite{zewde2026gptimage2} used X posts
to study an earlier release. Repeating its version-silent queries would
not reliably identify 2.5: in a launch-day pilot, \nSilentTwo{} of
\nSilentTotal{} returned posts explicitly named GPT-Image-2.
We therefore require explicit version evidence and retain its source.

Existing research already evaluates detection on social-media samples
and under online image processing~\cite{konstantinidou2026itwsm,
dellanna2025truefake,li2025rrdataset}. Here we examine a specific
collection problem raised by a product transition: what evidence supports
version attribution, and how do the resulting admission rules and source
mixture shape the observations? A companion paper studies forgery tasks
with known answers~\cite{images25forgery}; this paper concerns posted images
whose production histories are incompletely observed.

We organize the study around three questions:
\begin{enumerate}
  \item \textbf{What evidence supports version attribution?}
  We distinguish caption claims, caption claims with a creation-language
  classifier, and host model records. We document admission and curation
  decisions and inspect delivered content credentials
  (Sections~\ref{sec:gate}, \ref{sec:screen} and~\ref{sec:c2pa}).
  These records make the evidence auditable without treating every
  admitted image as independently authenticated.

  \item \textbf{How does source composition shape the collection?}
  We assemble \nImages{} images from \nPosts{} posts across
  \nPlatforms{} sources and report content and collection differences
  by source (Sections~\ref{sec:platforms} and~\ref{sec:content}).
  Query instruments, language metadata, admission rules and observed
  time coverage differ, so the source counts do not measure relative
  platform activity.

  \item \textbf{How do published detectors respond at fixed operating points?}
  We check six frozen implementations against a labelled benchmark,
  then report collection flag rates beside held-out control FPRs
  (Section~\ref{sec:detectors}).
  An exploratory X-only comparison with the April release extends this
  record across two collection windows (Section~\ref{sec:cross_version}).
  It does not isolate a generator-version effect.
\end{enumerate}

The resulting contribution is a documented collection of attributed
launch-period images and a fixed-threshold evaluation of detector responses
to it. Here, \emph{in the wild} means images retrieved from online posts and galleries
under the stated admission rules. The image-form filter excludes screenshots
and composites, and the collection is not a census of online use.
Historical audits assess agreement with another model; their coverage limits
are reported in Section~\ref{sec:audit}.

\section{Related Work}
\label{sec:related}

\paragraph{Model-specific datasets.}
Broad web-image collections such as LAION-5B~\cite{schuhmann2022laion} are not
organised around a specific generator. Targeted datasets focus on a generator or a defined
roster, as in DiffusionDB~\cite{wang2023diffusiondb} and
GenImage~\cite{genimage2023}. For OpenAI models,
GPT4o-Receipt~\cite{zhang2026gpt4oreceipt} collected document images with direct
API provenance. Our April dataset~\cite{zewde2026gptimage2} curated
\aprilImages{} GPT-Image-2 images from \nAprilImagePosts{} X posts under creator
self-report, starting from \aprilTweetsRead{} collected image records.
That collection is larger than the present one but uses a single platform and
attribution regime.

Two follow-up studies~\cite{textrich2026,detbench2026} predate Images~2.5, while
first-look studies of GPT-4o image
generation~\cite{yan2025gptimgeval,chen2025gpt4oimage} illustrate how to
characterise a newly released model. Our collection distinguishes explicit
caption-version claims from host-recorded attribution and applies an image-form
filter across multiple sources. The gateway in Section~\ref{sec:stack} reduces
the need for separate platform clients; it does not establish exhaustive coverage.

\paragraph{Evaluation in online settings.}
ITW-SM~\cite{konstantinidou2026itwsm} collects real and generated images from
four social platforms and evaluates how detector design and preprocessing
affect performance. TrueFake~\cite{dellanna2025truefake} studies generated
images shared through social networks, while RRDataset~\cite{li2025rrdataset}
examines scenario changes, internet transmission and re-digitization.
These studies already establish that online conditions matter for detection.
Our focus is attribution during a specific product transition: a tool can
retain its name while its underlying generator changes. We record the version
evidence available in captions and host fields, distinguish admission tiers,
and examine the resulting collection with fixed detectors. The April comparison
extends this observational record; it does not isolate the generator upgrade.

\paragraph{Provenance in delivered files.}
C2PA~\cite{c2pa2024} uses a signed manifest to describe an asset's production
history. OpenAI reports signing outputs and adding an invisible
watermark~\cite{openai2026systemcard}. In our April
study, embedded credentials did not survive the inspected X delivery path.
Here, the inspected generator-version field does not distinguish the tested
endpoint versions (Section~\ref{sec:c2pa}). X's ``Made with AI''
label~\cite{xlabels2024} and Pixiv's AI-generation declaration~\cite{pixivai} are additional
platform signals. Section~\ref{sec:limits} explains their limitations for this
collection.

\paragraph{Generated media and detection in practice.}
Self-reported AI images account for well under one percent of image posts in
the Reddit art communities studied by Matatov et al.~\cite{sha2024prevalence}.
That is a rate of disclosed AI use; without independently identifying undisclosed
AI images, it does not estimate disclosure recall.
Studies of AI political imagery on X~\cite{luo2025synthetic} and
fact-checked media~\cite{horton2024ammeba} document other settings in which
generated media circulate.

Image detectors often generalise poorly across generator
families~\cite{wang2020cnn,gragnaniello2021gan,ren2026howwell}. Multimodal
language models are inconsistent zero-shot detectors~\cite{ren2025mllm}, and
deployment studies find further performance losses outside laboratory
benchmarks~\cite{ren2025reality}. A collection attributed to a new generator,
with screenshots and composites filtered out, supports evaluation in this
less controlled setting. Its attribution and processing limits remain part of
that evaluation.

\section{Collection}
\label{sec:collection}

We first define which posts and assets enter the collection, then distinguish
version-attribution evidence from image form. These choices determine the
population described in the content and detector analyses.

\subsection{Launch anchor and funnel}
\label{sec:anchor}

\begin{figure*}[t]
  \centering
  \figph{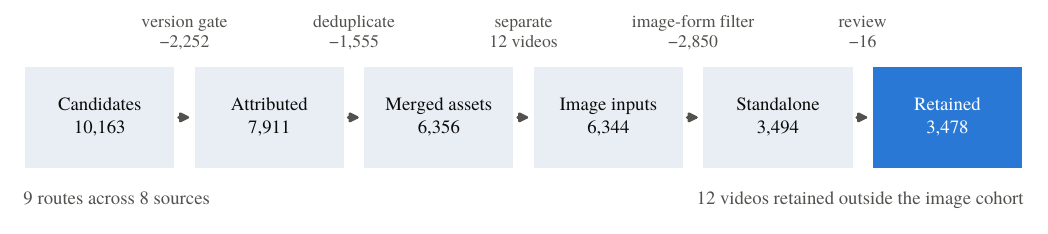}{1.0}
  \caption{Collection funnel. Of \nRecords{} candidate asset records,
  \nStrict{} passed caption or host-record attribution; deduplication removed \nDupKey{} exact
  (post, media, frame) repeats and \nDupImage{} average-hash near-duplicates
  (Appendix~\ref{app:dedup}), leaving \nMerged{} gated assets from
  \nMergedPosts{} posts. Image-form filtering kept the \nLabStandalone{} assets
  labelled standalone, and targeted review removed a further \nReviewedExcluded{},
  leaving \nImages{} retained images; \nVideo{} videos are
  recorded separately and excluded from image statistics.
  \texttt{dedup\_audit.py} replays the attribution and deduplication stages.}
  \label{fig:pipeline}
\end{figure*}

Every source is filtered against a single timestamp, the \mbox{@OpenAIDevs} announcement post,
whose snowflake identifier decodes to \launchStamp{}. Images~2.5 outputs existed
before it -- arena evaluations were live by \arenaDate{} -- but arena images are
anonymous and pre-launch API access was private, so the announcement is a reproducible
scope boundary rather than a claim about the earliest attributable generation. Official
X queries pass it as \texttt{start\_time}; other sources take the coarsest server-side
date filter they support and are filtered client-side against it
(Appendix~\ref{app:instruments}). Gated collection and released subset share an end,
\windowEnd{}, roughly \windowHours{} hours after the anchor. Figure~\ref{fig:pipeline}
traces the records from collection to the retained images.

\subsection{Collection infrastructure}
\label{sec:stack}

Five of the \nRoutesWord{} collection routes use one commercial gateway
(\gatewayVendor{}~\cite{monid2026}, CLI \gatewayCliVersion{}); Pixiv, note.com and NightCafe
use public JSON endpoints directly. The gateway log records \nGatewayCalls{}
calls across \nGatewayEndpoints{} endpoints. Endpoint details and cost
accounting appear in Appendix~\ref{app:gateway}.

The official and gateway X routes provide partly overlapping observations:
\nGatewayReturnedByOfficial{} of \nGatewayPosts{} gateway-admitted posts
appear in retained official results, and \nOfficialGatewayOverlap{} pass both
admission rules. Different queries, schedules and admission rules prevent
interpreting this difference as a controlled comparison of index coverage
(Appendix~\ref{app:overlap}).

\subsection{Sources, routes and yield}
\label{sec:platforms}

We collect X/Twitter through two routes. The official v2 recent-search
endpoint~\cite{xapi2024} has a seven-day retention window and does not provide
a census of public posts. A launch-day pilot and a day-one run used the same
queries. The day-one run read \nTweetsRead{} posts for \$\costOfficialDayOne{}
and yielded \nOfficialStrictRecords{} gate-passing records from
\nOfficialStrictTweets{} posts.

The gateway runs the same search but returns a partly overlapping set
(Appendix~\ref{app:overlap}). Both routes deliver media from the same content
delivery network at the same resolutions. We merge them on post and media
identifiers. The remaining seven sources use either the gateway
(Xiaohongshu, Weibo and Reddit through TikHub; Instagram through a resold Apify
hashtag scraper) or direct access (Pixiv, note.com and NightCafe).
Table~\ref{tab:platforms} gives retained image counts.

The \nRoutesWord{} instruments differ in query, time filtering, timestamp
resolution, served format, language metadata and creation-language requirements
(Appendix~\ref{app:instruments}). Cross-source differences therefore reflect
the instruments as well as the platforms. Reddit is mainly a discussion source
in this sample: \nNoImageRd{} of \nNamedTwoFiveRd{} qualifying posts have no
downloadable image. Instagram differs in being accessed by hashtag through a
third-party scraper rather than a search index (Appendix~\ref{app:sources}).

\begin{table}[t]
  \centering\small
  \caption{Retained images by platform after deduplication, image-form
  filtering and targeted review. X combines both collection routes. The separate \nVideo{} videos
  are excluded.}
  \label{tab:platforms}
  \setlength{\tabcolsep}{16pt}
  \begin{tabular}{lr}
    \toprule
    \textbf{Platform} & \textbf{Images} \\
    \midrule
    X/Twitter   & \nKeptTw{} \\
    NightCafe   & \nKeptNightcafe{} \\
    Xiaohongshu & \nKeptXhs{} \\
    note.com    & \nKeptNote{} \\
    Pixiv       & \nKeptPx{} \\
    Weibo       & \nKeptWb{} \\
    Instagram   & \nKeptIg{} \\
    Reddit      & \nKeptRd{} \\
    \midrule
    Total       & \nImages{} \\
    \bottomrule
  \end{tabular}
\end{table}

\subsection{The version gate}
\label{sec:gate}

We record three attribution tiers. Caption-based admission requires a
case-insensitive match to a 2.5 name, an API codename (Flare or Sunburst), or a
2.5 hashtag, with no accompanying 2.0 signal (Appendix~\ref{app:regex}). Posts
naming only 2.0, both generations, or neither are excluded from this route.
A match establishes a caption claim, not the attachment's true generator.

\begin{itemize}
  \item \textbf{Caption plus classifier} includes \nOfficialTier{} released
  images from official-API X records. Alongside the version match, these posts
  pass the April creation-language classifier. It uses creation phrases,
  shared prompts and showcase hashtags, while excluding specified announcement
  and promotional patterns.
  \item \textbf{Caption-only} includes \nCaptionTier{} images from gateway X
  and the other caption-based sources.
  \item \textbf{Host-recorded} includes \nHostTier{} NightCafe images admitted
  through the platform's generator field. They need no version-bearing caption.
  Its \nVideo{} videos form a separate cohort.
\end{itemize}
The tier field allows each population to be selected independently
(Appendix~\ref{app:tiers}). Host records state what the platform reports; we
did not independently verify endpoint execution.

The caption rule excludes version-silent posts and can also admit misleading or
non-creation captions. Its errors are not one-sided. The earlier text audit
assesses agreement about caption claims, with recall evidence restricted to
recorded official-X strata (Section~\ref{sec:audit}); it does not validate
host attribution or identify the true generator from pixels.

\subsection{Image-form filter and targeted review}
\label{sec:screen}

Attribution alone does not establish image form. A caption can refer to
Images~2.5 while its attachment is a screenshot; a host can identify the
generator while serving a grid or composite. We therefore classify each gated
image with \filterModel{} from a 768-pixel thumbnail, at a cost of
\$\costFilterPerK{} per thousand images.

The classifier assigns one of six image forms: a standalone image, chat or
app screenshot, collage or grid, photograph of a screen, promotional graphic,
or other. We retain only the standalone class. Appendix~\ref{app:prompt}
gives the exact labels and prompt. A targeted
visual review then removed \nReviewedExcluded{} prompt sheets, editorial
composites and end cards. Each exclusion has a recorded reason. This review
corrects identified errors but does not estimate the remaining error rate.

Of \nLabeled{} gated images, \nLabStandalone{} are labelled standalone and
\pctScreenRemoved\% are not: \nLabPromo{} promotional graphics, \nLabChatShot{} chat
screenshots, \nLabCollage{} collages or grids, \nLabScreenPhoto{} photographs of
screens and \nLabOther{} other. The filter separates image \emph{form} from
\emph{provenance}: the excluded collages and promotional graphics hold an unknown
number of real outputs in composite form, so exclusion is not a claim that an asset is
not AI-generated. Per-class counts by source are in Appendix~\ref{app:screentable}.

\subsection{Content credentials and version attribution}
\label{sec:c2pa}

A local byte scan found C2PA markers in \nPxCpaFound{} of \nPxCpaTotal{}
released Pixiv files; \nPxCpaNamesTwo{} contain the searched generator-version,
OpenAI and watermark strings. Marker presence is not signature validation, and
the remaining marker-positive file is not included in the matching-string claim
(Appendix~\ref{app:c2pa}). Earlier pilot signature checks are reported
separately. These observations do not establish preservation or stripping
behaviour for every source or delivery path.

The API-side ledger covers \nApiCpaTotal{} files generated for the benchmark on
\apiCpaDate{}: \nApiCpaFlare{} Flare, \nApiCpaSunburst{} Sunburst and
\nApiCpaTwo{} GPT-Image-2. All inspected files contain the generator-version
string \cpaGeneratorVersion{}, including both 2.5 endpoints. This is a census
of that logged file set, not of all API outputs. The supported result is that
the inspected generator-version field does not distinguish these tested endpoint
versions; caption and host attribution remain separate evidence sources.

\subsection{Coverage beyond the retained sources}
\label{sec:landscape}

The retained sources do not exhaust the available online images.
Launch-period probes also examined model-tagged galleries and API resellers
(Appendix~\ref{app:landscape}). A generation service need not expose a public
gallery, and a model filter need not return a model field for each image.

NightCafe illustrates the limits of a single probe. A \probeDateOne{}
sample of \nNightCafeSampled{} creations contained no 2.5 records.
On \probeDateTwo{}, the same endpoint yielded \nNcCreations{} records with
a 2.5 \texttt{gptImageModel} field. These samples came from a retrospectively
paginated feed. They describe observed feed contents, not when the platform
adopted the model. Empty results from other tested endpoints likewise establish
only what those queries returned. Collection records therefore retain the
endpoint, query and observation date as well as the attribution evidence.

\section{Historical Audits of Automated Curation}
\label{sec:audit}

We audited the version gate, screenshot filter and subject classifier on an
earlier collection, before the day-two expansion. Seeded samples
(seed \auditSeed{}) were relabelled by \auditorModel{}, a different model from
the screenshot filter's \filterModel{}. The audits measure agreement with another
model; they provide neither human annotation nor verified generator
attribution. They cost \$\auditCost{}.

The screenshot audit sampled a pool of \nAuditScreenPopulation{} images,
including \nAuditScreenPopStandalone{} predicted standalone images. Neither its
\nAuditScreen{} items nor the \nAuditTax{} subject-classification items include
NightCafe. These audits therefore do not validate the expanded release or its
host-attributed tier. Appendix~\ref{app:auditdetail} gives the sampling design.
The intervals below are descriptive Wilson 95\% intervals for raw sample
proportions, not uncertainty estimates adjusted for the weighted design.

\paragraph{Caption claims.}
Among \nAuditGateAdmitted{} sampled admitted post texts, raw agreement on the text
claim is \pctGatePrec\% (\pctGatePrecLo--\pctGatePrecHi). Disagreements concern
announcements, tutorials and marketing that name the version without claiming
the poster's own generation. Appendix~\ref{app:gateaudit} gives language
breakdowns and the limits of the recall analysis.

\paragraph{Image form.}
The auditor labelled \pctScreenPrecStandalone\%
(\pctScreenPrecStandaloneLo--\pctScreenPrecStandaloneHi) of the
\nAuditScreenStandalone{} predicted standalone images as standalone.
Overall raw agreement is \pctScreenAgreeRaw\%, with unweighted
$\kappa=\screenKappa{}$. The auditor treated some predicted collages and
promotional graphics as standalone. Because the \texttt{other} stratum was not
sampled, the audit does not identify full-pool recall
(Appendix~\ref{app:filteraudit}).

\paragraph{Subject labels.}
On \nAuditTax{} earlier released images, CLIP and the auditor agree on
\pctTaxAgree\% of labels ($\kappa=\taxKappa{}$). We therefore describe content
as a distribution of CLIP assignments, not validated subject prevalence.
Appendix~\ref{app:taxaudit} reports class breakdowns and their limitations.

\section{Content and Source Composition}
\label{sec:stats}

We next describe the retained images, keeping source-specific observations
separate from pooled summaries; Figure~\ref{fig:sample_grid} shows one
illustrative image per source. Content labels and post metadata are
measurements of this collection, not estimates of platform-wide use.

\subsection{Subject matter, text and faces}
\label{sec:content}

We apply the April study's three analyses to \contentN{} released images:
zero-shot CLIP subject classification, OCR and face detection.
Appendix~\ref{app:contentsettings} gives the settings and explains why the April
comparison is not controlled.

\begin{figure*}[t]
  \centering
  \figph{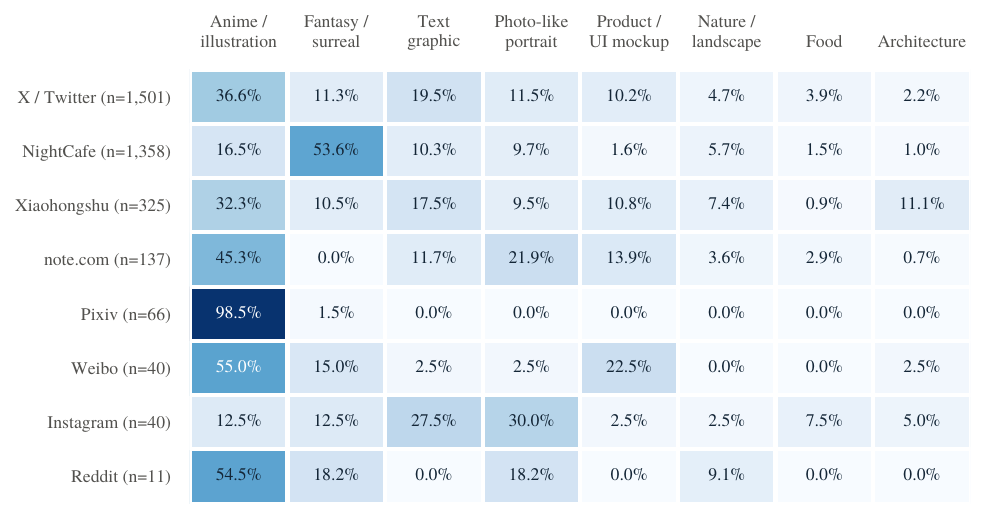}{1.0}
  \caption{CLIP-assigned subjects by source. Cells give percentages within each
  source, and row labels give image counts. The eight-class assignments show
  how source composition changes the aggregate distribution. They are not
  validated estimates of platform preferences or model capabilities.}
  \label{fig:platform_content}
\end{figure*}

Source composition strongly affects the pooled subject profile
(Figure~\ref{fig:platform_content}). NightCafe supplies \pctNcImageShare\% of
images but \pctNcFantasyShare\% of fantasy/surreal assignments. That class
accounts for \pctNcFantasy\% of NightCafe images and \pctTwFantasy\% of X
images. Removing NightCafe lowers the pooled fantasy share from
\pctClsFantasy\% to \pctFantasyWithoutNc\%. These differences describe the
sampled images. The small source cohorts shown in the figure cannot provide
stable estimates of platform-wide preferences.

OCR detects text in \pctText\% of images (\nText{}), compared with
\aprilPctText\% in April. The median number of text regions per image is
\medianTextRegions{} (April: \aprilMedianRegions{}). Face detection finds at
least one face in \pctFaces\% of images (\nFaces{}), compared with
\aprilPctFaces\% in April. These measurements characterise the collections;
they do not establish changes in model use. Because face detectors are biased
on non-photorealistic images and CLIP assigns much of this collection to
non-photographic classes, we
omit the demographic histograms used in the April paper.

\subsection{Languages, timing and wording}

The cohort contains \nImages{} images from \nPosts{} posts. Recorded language
labels span \nLangCountReal{} languages, but their sources differ.
For \nXLang{} X image rows, the label is the API's detected \texttt{lang}.
Another \nPlatformDefaultLang{} rows (\pctPlatformDefaultLang\%) use a platform
default, and \nLangNull{} rows (\pctLangNull\%) have no language field
(Appendix~\ref{app:lang}). These labels describe post metadata, not text
recognised in the images.

The most common recorded labels in the released cohort are Japanese
(\nLangJA{}), English (\nLangEN{}) and Chinese (\nLangZH{}). Chinese is most
common in the gated pool (\nGatedLangZH{}), but its retention rate is
\pctRetainLangZH\%, compared with \pctRetainLangJA\% for Japanese and
\pctRetainLangEN\% for English. Source composition and image-form filtering
both contribute to this difference; these counts do not measure platform adoption.

Recorded posting timestamps span roughly \windowHours{} hours and differ from
retrieval dates (Appendix~\ref{app:timing}). Of \nLateImages{} retained images
posted from \ncWindowStart{} onwards, \nLateNightCafe{}
(\pctLateNightCafe\%) come from NightCafe and were collected on
\probeDateTwo{}. The final interval therefore reflects that source's collection
route rather than demonstrating an increase in posting activity.

We analyse wording in \nWordEligiblePosts{} caption-attributed posts with
nonempty text. NightCafe is excluded because its \texttt{text} field contains
creation titles rather than caption claims. English, Japanese and Chinese
patterns match prompt vocabulary in \pctWordPrompt\% of eligible posts,
editing in \pctWordEdit\%, consistency in \pctWordConsistency\% and speed in
\pctWordSpeed\%. Sunburst and Flare appear in \pctWordSunburst\% and
\pctWordFlare\%, respectively. These unvalidated keyword matches do not measure
how often users edited images or disclosed prompts.

\section{Detector Responses to the Collection}
\label{sec:detectors}

We evaluate six frozen image detectors: Community Forensics~\cite{cf2025},
B-Free~\cite{bfree2025}, Effort~\cite{effort2025}, PGC~\cite{zhou2026pgc}
(2026, SD~v1.4 checkpoint), PROBE-ResNet50~\cite{cao2026probe}, and
DoU~\cite{he2026dou}. None is trained or fine-tuned on this collection.
Because generator identity is creator- or host-attributed, the outcome is the
fraction of images \emph{flagged}, not verified detection recall.

\subsection{Controls and operating point}

We select 400 reference images from each of five photograph pools: GeoDE,
SUN397, FairFace, Pascal VOC and Food101. We separately select 400 WikiArt
artworks. Source labels are inherited without new human adjudication.
After exact-byte deduplication, deterministic hash ordering and alternating
assignment produce equal calibration and test splits within each pool.

Each detector's native-logit threshold is calibrated on the 1,000 calibration
photographs to flag at most 5\%. The 1,000 held-out photographs measure the
realised false-positive rate (FPR). Artwork controls provide a separate
diagnostic and do not set the thresholds. Every detector scores all 3,478
collection images and 2,400 controls without decoding or inference failures.
DoU retains its stochastic evaluation forward pass with primary seed 42;
Appendix~\ref{app:detectors} gives execution details and a second-seed check.

\paragraph{What the operating point establishes.}
Photograph calibration fixes a reproducible threshold for each detector.
It does not establish a 5\% FPR on every source or content type.
The held-out controls measure how often reference photographs and artworks
are incorrectly flagged; the collection rate measures how often attributed
images are flagged. Deployment precision would additionally require
representative controls and the prevalence of generated images in the target
stream. These experiments therefore support comparisons of the frozen
cohorts, not a deployment ranking.

\begin{table*}[htbp]
\centering
\caption{Frozen-detector results at a 5\% reference-photograph calibration FPR.
Collection intervals are 95\% post-clustered bootstrap intervals conditional
on the fitted threshold. Held-out control FPRs are separated by content type.
Collection flag rates are not independently verified recall. DoU uses seed 42;
its intervals exclude latent-sampling variation.}
\label{tab:detectors}
\begin{tabular}{lrrr}
\toprule
Detector & Collection flagged & Photo FPR & Artwork FPR \\
 & ($n=3{,}478$) & ($n=1{,}000$) & ($n=200$) \\
\midrule
Community Forensics & 54.7\% [52.8, 56.7] & 5.9\% & 42.0\% \\
B-Free & 15.0\% [13.4, 16.6] & 4.2\% & 64.0\% \\
Effort & 43.7\% [41.5, 46.2] & 3.5\% & 8.5\% \\
PGC (2026) & 56.4\% [54.2, 58.6] & 5.1\% & 96.5\% \\
PROBE-ResNet50 & 3.7\% [2.6, 5.1] & 5.3\% & 1.5\% \\
DoU & 17.4\% [15.1, 19.5] & 4.4\% & 11.0\% \\
\bottomrule
\end{tabular}

\end{table*}

\subsection{Harness reproduction on GenImage}
\label{sec:reproduction}

We check the six detector implementations on a labelled benchmark before
interpreting their collection scores. Forward-pass agreement with official
code and plausible photograph FPRs are insufficient: the former can share a
preprocessing error, and the latter tests no generated images.

GenImage~\cite{genimage2023} provides a common reference. Community Forensics
reports results on it, Effort's released checkpoint targets it, and PROBE and
PGC train on its SD~v1.4 split. We score a frozen, balanced validation sample:
\nGenImagePerClass{} real and \nGenImagePerClass{} generated images per
generator across \nGenImageGenerators{} generators, totalling
\nGenImageImages{} images. The sample is hash-selected before inference, with
no fitting or tuning. This implementation check uses each detector's own
$0.5$ boundary. The collection comparison in Section~\ref{sec:flagrates}
instead applies the photograph-calibrated thresholds to both GenImage and
the collected images. Appendix~\ref{app:reproduction} gives the freeze,
per-generator comparisons and a separate Community Forensics evaluation.

\begin{table*}[htbp]
\centering\small
\caption{Accuracy on a balanced \nGenImageImages{}-image GenImage validation
sample at each detector's own $0.5$ boundary. Means use the same
\nGenImageGenerators{} generators; $\Delta$ is ours minus published.
$^{\dagger}$Community Forensics reports pooled accuracy over a set containing
one additional generator, so its comparison is not exactly matched.
$^{\ddagger}$B-Free's reference is another group's measurement.
DoU reports GenImage results only graphically, so no tabulated reference is
available.}
\label{tab:reproduction}
\begin{tabular}{lrrrr}
\toprule
Detector & Acc. (\%) & AP (\%) & Published (\%) & $\Delta$ \\
\midrule
Community Forensics & 95.0 & 99.8 & 95.7\rlap{$^{\dagger}$} & $-$0.7 \\
B-Free & 87.2 & 96.9 & 87.3\rlap{$^{\ddagger}$} & $-$0.1 \\
Effort & 92.1 & 98.3 & 88.7 & +3.4 \\
PGC (2026) & 99.7 & 100.0 & 98.1 & +1.6 \\
PROBE-ResNet50 & 74.4 & 86.2 & 75.5 & $-$1.1 \\
DoU & 88.7 & 99.8 & not reported & -- \\
\bottomrule
\end{tabular}

\end{table*}

All five detectors with a published reference agree within \gimMaxDelta{}
accuracy points, and four agree within \gimTightDelta{}
(Table~\ref{tab:reproduction}). Several per-generator patterns also reproduce.
PROBE-ResNet50 reaches \gimProbeAdmOurs\% versus the published
\gimProbeAdmPub\% on ADM, and \gimProbeBigganOurs\% versus
\gimProbeBigganPub\% on BigGAN. B-Free agrees within \gimBfreeMaxDiff{}
points on all seven generators.

PGC's published accuracy is 100.0\% on \gimPgcSaturated{} generators.
BigGAN is its only nonsaturated reference cell and the one that disagrees.
Across all \gimNComparisons{} per-generator comparisons, three differences
exceed three standard errors of our subset estimate, all with our result
higher. The remainder lie within three standard errors, in both directions.

The benchmark results provide evidence against inverted scores or gross
implementation failures. They do not prove every collection preprocessing
path correct. PROBE-ResNet50's reproduced benchmark weaknesses are consistent
with its low collection flag rate of \pctProbeWild\%, although that consistency
alone does not identify the cause of the collection result.

\subsection{Flag rates on the collection}
\label{sec:flagrates}

Held-out photograph FPRs range from \pctFprLow{} to \pctFprHigh\%, while
artwork FPRs range from \pctArtLow{} to \pctArtHigh\%
(Table~\ref{tab:detectors}). PGC, for example, flags \pctPgcArt\% of reference
artworks and \pctPgcPhoto\% of photographs. A high collection flag rate can
therefore coexist with sensitivity to human artwork.

At the same thresholds, the six detectors recall
\pctGenRecallLow--\pctGenRecallHigh\% of GenImage's generated images with at
most \pctGenFprHigh\% false positives. They flag
\pctWildFlagLow--\pctWildFlagHigh\% of the collection. For every detector,
the latter rate is lower, by \ppGapLow{}--\ppGapHigh{} percentage points.

\begin{figure*}[t]
\centering
\includegraphics[width=\textwidth]{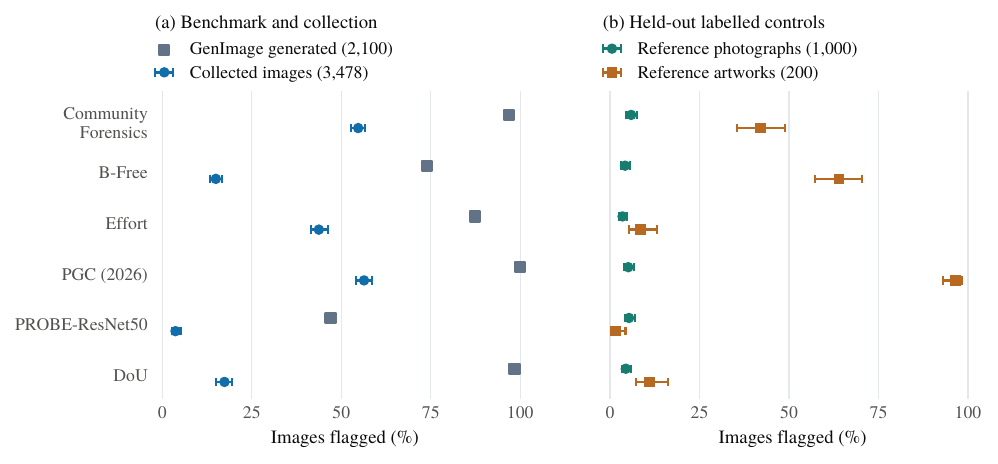}
\caption{Detector responses at the same photograph-calibrated thresholds.
Left: labelled GenImage generated-image recall (squares) and attributed
collection flag rates (circles). Right: false-positive rates on held-out
reference photographs and artworks. Collection bars are 95\% post-clustered
bootstrap intervals; control bars are 95\% Wilson intervals. GenImage points
show the frozen sample estimates. Sample sizes appear in the legends.
The populations differ in content, provenance and processing; connecting
their outcomes to a generator effect would require a controlled experiment.}
\label{fig:detector_overview}
\end{figure*}

Figure~\ref{fig:detector_overview} places these rates beside the held-out
photograph and artwork controls. It makes both observations visible: collection
flag rates are low relative to benchmark recall, while artwork FPRs can be
high. Interpreting either requires the corresponding population and denominator.

The benchmark check supports the implementation for
\nGimSelfReported{} detectors with their own published reference results
(Section~\ref{sec:reproduction}). It does not exclude preprocessing faults
specific to platform-recompressed images: GenImage uses small benchmark PNGs.
Three per-generator comparisons exceed sampling error, all with our accuracy
above the published value. Those discrepancies widen the observed gap.

\paragraph{Interpreting the gap.}
Five additional differences prevent treating this gap as an isolated measure
of detector generalization:
\begin{enumerate}
  \item \textbf{Attribution is unverified.} Images incorrectly attributed to
  Images~2.5 can change the collection flag rate.
  \item \textbf{Processing differs.} Some delivery routes recompress images, and we have
  not bounded the resulting effect on detector scores.
  \item \textbf{Content differs.} GenImage is photographic and ImageNet-derived;
  CLIP assigns much of this collection to non-photographic classes. The artwork controls show changes
  of tens of percentage points between the two control types; their content
  and processing are not matched.
  \item \textbf{Benchmark exposure differs.} Effort's checkpoint is selected
  for GenImage, while PROBE and PGC train on its SD~v1.4 split. Removing the
  two SD-family generators changes the gap range to
  \ppGapOodLow{}--\ppGapOodHigh{}~pp, mostly through PROBE-ResNet50.
  \item \textbf{Some images may be edits.} Editing vocabulary appears in
  \pctWordEdit\% of \nWordEligiblePosts{} caption-attributed posts. These
  unvalidated keyword matches exclude NightCafe and are measured per post,
  so they cannot estimate or bound the share of edited images. Edits may
  retain real reference content and need not resemble wholly generated
  benchmark images.
\end{enumerate}
We report the observed gaps without assigning their magnitude to any one cause.

\paragraph{Attribution and source sensitivity.}
Restricting the collection to its most restrictive caption-based tier does not narrow the
gap consistently. Among \nOfficialTier{} official-X images admitted by both
caption matching and the creation classifier, the gap widens for
\nStrictTierWidens{} of six detectors and ranges from
\ppGapStrictLow{} to \ppGapStrictHigh{}~pp. The gaps therefore persist in this
subset. The restriction does not validate attribution or isolate its effect:
the tier is X-only, so tier and platform remain confounded.

Flag rates also vary by source (Appendix Table~\ref{tab:detector_sources}).
Content, attribution and processing differences are intertwined in those
comparisons. These results show why source and
attribution fields are needed when reusing the dataset for detection studies.

\subsection{Comparison with our April GPT-Image-2 release}
\label{sec:cross_version}

\begin{table*}[htbp]
\centering\small
\caption{X-only flag rates at identical detector-specific thresholds.
Entries are percentages with 95\% post-clustered bootstrap intervals;
differences are September minus April in percentage points (pp). April uses
\nAprilSample{} of \aprilImages{} released images; September X contains
\nXImages{}. The contrasts do not identify causal generator-version effects.
DoU uses seed 42.}
\label{tab:cross_version}
\begin{tabular}{lrrr}
\toprule
Detector & April X (\%) & September X (\%) & Difference (pp) \\
\midrule
Community Forensics & 44.8 [42.9, 46.7] & 44.3 [41.2, 47.5] & $-$0.5 [$-$4.1, +3.2] \\
B-Free & 18.1 [16.6, 19.5] & 19.3 [16.8, 22.0] & +1.3 [$-$1.7, +4.3] \\
Effort & 22.7 [21.1, 24.3] & 28.8 [25.8, 31.9] & +6.1 [+2.7, +9.6] \\
PGC (2026) & 42.4 [40.5, 44.2] & 43.2 [39.8, 46.7] & +0.8 [$-$3.1, +4.7] \\
PROBE-ResNet50 & 2.9 [2.3, 3.5] & 3.8 [2.8, 4.9] & +0.9 [$-$0.3, +2.2] \\
DoU & 7.6 [6.6, 8.6] & 11.1 [8.8, 13.6] & +3.6 [+1.0, +6.2] \\
\bottomrule
\end{tabular}

\end{table*}

\paragraph{Design.}
We score a fixed \nAprilSample{}-image sample from \nAprilSamplePosts{} posts
in the April GPT-Image-2 release~\cite{zewde2026gptimage2}. Its parent release
contains \aprilImages{} images from \nAprilImagePosts{} X posts.
We use the same six checkpoints, preprocessing and score definitions as for
September, whose scores are already complete. A deterministic hash ranking
selects April images before inference. Each detector retains its
photograph-calibration threshold; neither cohort sets the operating point.

The primary September cohort comprises \nXImages{} X images from
\nXPosts{} posts. The pooled release is a secondary view. The cohorts are
unpaired, and 95\% percentile intervals use \xvDraws{} post-clustered bootstrap
draws per cohort. A 100-image pilot from the same ranking found no separation
and resolved differences only to about $\pm$10~pp. We then extended the prefix.
This added records without reselecting them, but the expansion decision
followed the pilot results. The reported intervals do not account for that
look, so these contrasts are exploratory. A held-out April check appears below;
Appendix~\ref{app:cross_version} gives coverage and sensitivity details.

\paragraph{Comparability.}
April admitted generic AI-badge and name-only claims without an image-form
filter. Thus, even within X, the cohorts differ in attribution rules,
screenshots, content and collection window. Detector-training exposure to either
release is unverified. This design compares two collections and cannot isolate
the effect of upgrading the generator.

\begin{figure}[htbp]
\centering
\includegraphics[width=\linewidth]{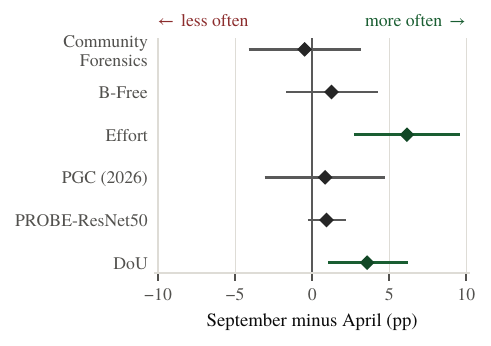}
\caption{Flag-rate differences between the September and April collections
at fixed detector-specific thresholds. Positive values mean September X images
are flagged more often. Bars are pointwise 95\% post-clustered bootstrap intervals; those excluding zero are
green. They omit DoU's latent-sampling variability. Per-collection rates appear
in Table~\ref{tab:cross_version}.}
\label{fig:cross_version}
\end{figure}

\paragraph{Findings.}
The fixed-threshold 95\% intervals indicate higher September X flag rates
for Effort and DoU (Table~\ref{tab:cross_version}, Figure~\ref{fig:cross_version}).
The differences are \xvEffortDiff{}~pp
[\xvEffortLo{},~\xvEffortHi{}] for Effort and \xvDouDiff{}~pp
[\xvDouLo{},~\xvDouHi{}] for DoU. Both remain positive after a Bonferroni
adjustment for \xvNDetectors{} contrasts: Effort
[\xvEffortBonfLo{},~\xvEffortBonfHi{}] and DoU
[\xvDouBonfLo{},~\xvDouBonfHi{}]. We use \xvDraws{} bootstrap draws to
estimate these tail bounds.

DoU's adjusted interval only narrowly excludes zero and omits its
latent-sampling variability. A second seed shifts the pooled September flag
rate by \pctDouSeedShift{}~pp at the same threshold. We therefore treat its
contrast as suggestive. The other four detectors show no detected difference,
with interval half-widths at most \xvMaxHalfWidth{}~pp. These intervals quantify
precision but do not establish equivalence.

No detector's interval lies wholly below zero. All point estimates except
Community Forensics' are positive. These are differences between the sampled
X collections; they do not establish whether the upgraded generator is
intrinsically easier or harder to detect.

\paragraph{Check with held-out April images.}
We re-test the two positive contrasts on \nHoldoutImages{} April images from
\nHoldoutPosts{} posts at ranks beyond the analysed sample. These images
were not used in the sample-expansion decision. Thresholds, score definitions
and September scores remain fixed, and only Effort and DoU are evaluated.

Effort's held-out contrast is \xvEffortHoldDiff{}~pp
[\xvEffortHoldLo{},~\xvEffortHoldHi{}].\holdoutDou{}
Both contrasts remain positive under the same family-wise adjustment, with
lower bounds of \xvEffortHoldBonfLo{} and \xvDouHoldBonfLo{}~pp.
Only April is held out. Reusing the same \nXImages{} September images makes
the estimates correlated, so this check replicates the April side of the
comparison rather than the full contrast independently. Because selection is
by image rank, \nHoldoutSharedPosts{} posts contribute images to both April
samples, including \nHoldoutSharedImages{} held-out images; the held-out images
are therefore not independent of the analysed sample at the post level.

\paragraph{Calibration uncertainty.}
A further post hoc analysis resamples the calibration photographs within
their five source pools, refits each threshold, and resamples posts in both
cohorts. The six-detector Bonferroni intervals remain positive for Effort
[\calEffortBonfLo{},~\calEffortBonfHi{}] and DoU
[\calDouBonfLo{},~\calDouBonfHi{}]~pp. The other four include zero
(Appendix Table~\ref{tab:calibration_sensitivity}). This check propagates
calibration sampling but still omits DoU's latent-sampling variability.

\paragraph{Composition sensitivity.}
Within September, widening from X to the pooled collection changes flag rates
by up to 14.9~pp, in both directions. For Community Forensics, Effort, PGC and
DoU, this shift exceeds every April--September difference in
Table~\ref{tab:cross_version}. Narrowing both cohorts to metadata-based
attribution subsets (April \texttt{confirmed} images and September official-X
classifier admissions) leaves the Effort and DoU differences positive; only
PROBE-ResNet50, whose X difference is near zero, changes sign. Only the X-only
comparison has a difference interval; pooled and attribution-subset views are
descriptive.

\section{Limitations}
\label{sec:limits}

\paragraph{Exploratory detector evaluation.}
Collection labels record attribution. Control labels are inherited from
existing datasets, without matching their content or processing to the
collected images. Detector-training overlap and residual perceptual duplicates
are unverified, and six checkpoints cover only part of the detector landscape.
Primary intervals condition on the calibrated thresholds; an additional
sensitivity analysis resamples calibration photographs. DoU also samples latent noise:
we fix its seed, and the reported intervals exclude that source of variation.

\paragraph{Scope of the implementation checks.}
For the five detectors with a published GenImage reference, our accuracies
agree within \gimMaxDelta{} percentage points; DoU has no tabulated reference. This agreement provides evidence against gross scoring
errors, such as inverted logits, on that benchmark. It cannot distinguish a
harness fault from a difference between the released checkpoint and the one
used for the paper's table. Nor does it validate preprocessing for the
platform-processed collection. B-Free's reference comes from another group.

\paragraph{Unmatched cross-release cohorts.}
The April and September cohorts are not matched by image, prompt or creator.
Exact-byte and decoded-pixel checks found no shared images; creator and prompt
overlap was not established. Even within X, the cohorts differ
in admission rules, image-form filtering and collection window as well as
attributed generator version. The pooled September view additionally changes
the platform mix. The \xvNSignificant{}
contrasts excluding zero survive a Bonferroni adjustment for
\xvNDetectors{} detectors, but they identify collection differences.
The observed source shifts are large enough to be plausible alternative
explanations. Effort has both the largest cross-release difference and the
largest deviations from its published per-generator results. Content-dependent
implementation differences would not necessarily cancel between these cohorts.

Appendix~\ref{app:limitations} details the collection's remaining limitations:
attribution, historical audit coverage, selection, source comparability,
near-duplicate removal and snapshot coverage. Platform AI labels likewise do
not resolve generator version or establish coverage of all relevant posts.

\section{Conclusion and Release}
\label{sec:conclusion}

This study documents the attribution evidence available when collecting
online images after an image tool is upgraded without a new public name.
The frozen snapshot contains \nImages{} images from \nPosts{} posts
across \nPlatforms{} sources, plus \nVideo{} separately labelled videos.
Caption claims and host records provide different admission evidence,
and the inspected C2PA generator-version field does not distinguish
the tested 2.5 endpoints from GPT-Image-2. None of these observations
independently authenticates every image's production history.

Observed content profiles and detector flag rates differ across sources.
These results motivate retaining source and attribution fields
when reusing the collection. They do not estimate platform-wide preferences
or activity. Historical model-based audits leave a need for human review
of the frozen cohort's evidence and curation decisions; visual review alone
cannot authenticate a generator version.

At fixed thresholds, collection flag rates are lower than GenImage recall
for all six detectors. The gaps remain compatible with attribution, content and processing differences.
In the exploratory April--September X comparison, fixed-threshold intervals
support a higher September flag rate for Effort and suggest one for DoU.
Both contrasts persist with held-out April images, which share posts with the
analysed sample, and with calibration resampling; DoU's latent-sampling
variation remains excluded.
Matched prompts, verified generator records and controlled image processing
would be needed to identify an upgrade effect.

\paragraph{Prepared artifacts and release.}
The local snapshot contains media, \texttt{metadata.csv} and a datasheet
(Appendix~\ref{app:release}). The collection and curation code records
queries, endpoints, gate expressions, deduplication and the screenshot
prompt. The dataset and code are released at
\mbox{\url{https://scam.ai/research}}.

\section*{Data and Code Availability}
The datasets and code supporting this work are released by Scam.AI and are available from
the Scam.AI Research portal, \url{https://scam.ai/research}~\cite{scamai_research}. Document-
and PDF-forgery resources are additionally maintained at Trustable, \url{https://trustable.ai}~\cite{trustable_ai}.
These are the canonical, maintained locations for the resources associated with this paper and
supersede any prior ad hoc distribution.

{\small
\let\originalbibitem\bibitem
\renewcommand{\bibitem}{%
  \ifnum\value{enumiv}=22\relax\balance\fi
  \originalbibitem}
\bibliographystyle{plain}
\bibliography{references}
}

\clearpage
\nobalance
\appendix
\section*{Appendix}
\addcontentsline{toc}{section}{Appendix}

\noindent The appendix provides four groups of supporting material:
\begin{itemize}
  \item \textbf{Collection design:} launch context (\ref{app:arena}), gateway
  accounting (\ref{app:gateway}), differences among \nRoutesWord{}
  instruments (\ref{app:instruments}), route overlap (\ref{app:overlap}),
  queries (\ref{app:queries}) and attribution rules (\ref{app:regex}).
  \item \textbf{Curation and audits:} deduplication (\ref{app:dedup}), the
  screenshot prompt (\ref{app:prompt}) and historical audits (\ref{app:auditdetail}).
  \item \textbf{Dataset evidence:} languages, timing and geometry
  (\ref{app:lang}), content credentials (\ref{app:c2pa}), the source landscape
  (\ref{app:landscape}), prepared release files (\ref{app:release}) and
  aggregate subject shares (\ref{app:taxonomy}).
  \item \textbf{Evaluation:} detector execution, benchmark checks,
  cross-release comparisons and collection limitations (\ref{app:detectors}).
\end{itemize}

\section{Model context at launch}
\label{app:arena}

On the Arena.ai blind-vote leaderboards snapshotted on \arenaDate{}, a day
before the announcement, both 2.5 models sat above GPT-Image-2 on editing
(\eloEditSunburst$\pm$\ciEditSunburst{} and \eloEditFlare$\pm$\ciEditFlare{}
against \eloEditTwo$\pm$\ciEditTwo{} Elo, on \votesEditSunburst{},
\votesEditFlare{} and \votesEditTwo{} votes, for Sunburst, Flare and GPT-Image-2
respectively). On text-to-image, both also sat above GPT-Image-2, Flare by a
narrow margin (\eloTtiSunburst$\pm$\ciTtiSunburst{} and
\eloTtiFlare$\pm$\ciTtiFlare{} against \eloTtiTwo$\pm$\ciTtiTwo{}). The system
card~\cite{openai2026systemcard} reported a lower unsafe-presented rate
(\unsafeSunburst\% and \unsafeFlare\% against \unsafeTwo\%).

\section{The billed collection stack}
\label{app:gateway}

The April collection used one scraper on one platform. Extending it required
additional clients, credentials, pagination logic and rate-limit handling.
Here, five of \nRoutesWord{} routes share a commercial gateway
(\gatewayVendor{}~\cite{monid2026}, CLI \gatewayCliVersion{}). Each call specifies a provider,
endpoint and JSON query, reducing the client code needed for another platform.

The logs contain \nGatewayCalls{} calls across \nGatewayEndpoints{}
endpoints, with latency and return code recorded per call.
Table~\ref{tab:gateway} lists counts and prices. Pixiv, note.com and NightCafe
instead use public JSON endpoints directly, without a gateway charge.

\paragraph{Unit economics.} The official X
recent-search endpoint bills \$\priceOfficialTweet{} per post read; the same
search through the gateway bills \$\priceGatewayPage{} for a page of twenty
posts, about \gatewayPriceRatio$\times$ cheaper per post. Xiaohongshu is
\$\priceXhsPage{} per twenty notes and Weibo and Reddit \$\priceGatewayPage{}
per page. These are list-price ratios for fully populated pages, before repeated posts,
failed calls and deduplication; they are not ratios per unique released image.
The first multi-platform crawl exhausted its \$\costGateway{} balance, and
Reddit and Instagram were added after a top-up (observed balance
\$\balanceAfterTopUp{}). The official day-one X run cost
\$\costOfficialDayOne{} for \nTweetsRead{} posts. These costs reflect the recorded call schedule and may differ for
another collector.

\paragraph{Observed route overlap.} Of \nGatewayPosts{} gateway-admitted
posts, \nGatewayReturnedByOfficial{} appear in the retained official result
files; \nOfficialGatewayOverlap{} pass both admission rules, and
\nGatewayReturnedFailedGate{} were returned officially but rejected there.
The remaining \nGatewayNotObservedOfficial{} gateway-admitted posts were not observed in the retained
official results. Those files do not preserve every billed read, and query
schedules and admission rules differ (Appendix~\ref{app:overlap}). These counts
therefore describe recorded overlap, not a causal decomposition of index
coverage.

Combining routes increases observed coverage. Reproduction still depends on the
gateway's prices, endpoint availability and retrieval behaviour. Direct public
endpoints avoid that intermediary where they are available.

The gateway calls in Table~\ref{tab:gateway} were logged by
\texttt{crawl/crawl\_all.py} and support Section~\ref{sec:stack}.

\begin{table*}[h]
  \centering\footnotesize
  \caption{The billed collection stack, from \texttt{data/crawl/crawl\_log.jsonl}:
  every gateway call, its endpoint, its measured mean wall-clock latency and the
  vendor's list price. Pixiv, note.com and NightCafe are absent: they were collected
  directly, without gateway charges.}
  \label{tab:gateway}
  \setlength{\tabcolsep}{3.5pt}
  \begin{tabular}{llrrl}
    \toprule
    \textbf{Source} & \textbf{Provider and endpoint} & \textbf{Calls} & \textbf{Mean s} & \textbf{List price} \\
    \midrule
    X/Twitter   & tikhub \texttt{/twitter/web/fetch\_search\_timeline} & \nCallsTwGateway{} & \secsTwGateway{} & \$\priceGatewayPage{}/20 posts \\
    Reddit      & tikhub \texttt{/reddit/app/fetch\_dynamic\_search}   & \nCallsRd{} & \secsRd{} & \$\priceGatewayPage{}/page \\
    Xiaohongshu & tikhub \texttt{/xiaohongshu/app\_v2/search\_notes}   & \nCallsXhs{} & \secsXhs{} & \$\priceXhsPage{}/20 notes \\
    Weibo       & tikhub \texttt{/weibo/web\_v2/fetch\_realtime\_search} & \nCallsWb{} & \secsWb{} & \$\priceGatewayPage{}/page \\
    Instagram   & apify \texttt{/apify/instagram-hashtag-scraper}      & \nCallsIg{} & \secsIg{} & per actor run \\
    \midrule
    Total       &                                                     & \nGatewayCalls{} & & \\
    \bottomrule
  \end{tabular}
\end{table*}

\section{What differs between the \nRoutesWord{} instruments}
\label{app:instruments}

Table~\ref{tab:instruments} compares the routes, including query type, time
filtering, media format and language metadata. These differences limit
cross-platform comparisons even when the launch cutoff is shared.

\begin{table*}[h]
  \centering\footnotesize
  \caption{The launch-time cutoff, deduplication and image-form filter are shared.
  Caption and host attribution use different admission rules; query instrument, time filter, timestamp
  resolution, served format, language field and creation-language requirement
  are not, so cross-source comparisons in the body are comparisons of
  instruments as much as of platforms. A hashtag scraper and a model-tagged gallery have different sampling
  frames from full-text search.}
  \label{tab:instruments}
  \setlength{\tabcolsep}{3pt}
  \begin{tabular}{lllllll}
    \toprule
    \textbf{Source} & \textbf{Query} & \textbf{Server time filter} & \textbf{Timestamp} & \textbf{Served as} & \textbf{Language} & \textbf{Creation lang.} \\
    \midrule
    X, official  & 6 families, full text & \texttt{start\_time} = anchor & second  & resampled JPEG      & API \texttt{lang}  & required \\
    X, gateway   & 6 strings, full text  & \texttt{since:} day           & second  & resampled JPEG      & API \texttt{lang}  & not required \\
    Xiaohongshu  & 4 strings, note search & one-week filter              & second  & transcoded webp     & constant \texttt{zh} & not required \\
    Weibo        & keyword, realtime search & none                      & minute, relative & CDN JPEG   & constant \texttt{zh} & not required \\
    note.com     & 4 hashtags            & none (newest first)           & second  & re-encoded by note  & constant \texttt{ja} & not required \\
    Pixiv        & 5 strings, caption search & none (newest first)       & second  & original bytes      & constant \texttt{ja} & not required \\
    Reddit      & 4 strings, post search & one-week filter              & second  & Reddit CDN JPEG     & none (no field)      & not required \\
    Instagram   & 3 hashtags + 1 keyword & none (scraper walk)          & second  & Instagram CDN JPEG  & none (no field)      & not required \\
    NightCafe   & newest creations feed & local anchor cutoff & millisecond & CDN assets & none (no field) & host model field \\
    \bottomrule
  \end{tabular}
\end{table*}

\section{The two X routes, and the two sources unlike the rest}
\label{app:overlap}

The retained official results contain \nGatewayReturnedByOfficial{} of the
\nGatewayPosts{} gateway-admitted posts, or
\pctGatewayReturnedByOfficial\%. This is a lower bound on returned overlap:
the curated files retain \nOfficialReturnedPosts{} posts with media, not all
\nTweetsRead{} billed reads.

The official creation-language classifier rejects
\nGatewayReturnedFailedGate{} of those returned posts, including
\nGatewayReturnedFailedGateRegex{} that pass the version regex alone
(Section~\ref{sec:gate}). Both routes admit \nOfficialGatewayOverlap{}
posts: \pctOfficialInGateway\% of \nOfficialStrictPosts{} official posts
and \pctGatewayInOfficial\% of gateway posts.

A further \nGatewayNotObservedOfficial{} gateway-admitted posts do not appear
in the retained official files. Thus, the records distinguish
\nGatewayReturnedFailedGate{} returned-but-rejected posts from
\nGatewayNotObservedOfficial{} unobserved posts. They cannot uniquely assign
unmatched posts to index coverage, because retention is incomplete and query
schedules differ. Neither route is a census. All \nPilotStrictPosts{}
gate-passing pilot posts also occur in the day-one run.

\label{app:sources}
Reddit mainly contributes discussion in this sample. Across four queries, the
crawl read \nFetchedRd{} posts, of which \nPostLaunchRd{} were post-launch
and \nNamedTwoFiveRd{} named 2.5. Only \nCollectorKeptRd{} had a directly
downloadable image. The other \nNoImageRd{} (\pctNoImageRd\%) were link
posts, workflow write-ups or excluded \texttt{preview.redd.it} re-encodes.
The \nKeptRd{} retained images document this route's yield but do not provide
a comparable platform sample.

On Instagram, \nNamedTwoFiveIg{} of \nFetchedIg{} returned posts named 2.5,
and all of those carried images. Access through a third-party hashtag scraper,
however, gives it a different sampling frame from full-text search.

\section{Query families, dropped families and animated posts}
\label{app:queries}

The official-API query set descends from the six April families, adapted for a
name with a decimal point: hashtags cannot contain a dot, so the community
settled on compressed forms such as \texttt{\#GPTImage25}, and the dotted name
appears only as phrase-quoted free text. Four April families were kept
(hashtags; English creation language with a name variant; Japanese creation
signals on ``generated with'' and ``tried making''; Chinese signals on
``prompt'' and ``generated''), two were dropped, and two were added: a codename
family targeting Flare and Sunburst, minus announcement phrasing, because those
tokens are dominated by third-party platforms announcing availability, and a
broad anti-noise family matching name variants minus comparison and release
language.

The gateway and non-X sources use shorter query strings
(Appendix~\ref{app:instruments}); the collection code records every string.

The two families dropped after the pilot are the feature-name queries, which
carry no version signal, and the version-silent family pairing ChatGPT with
prompt-sharing language---April's largest source of images---which is unusable
for a silently upgraded model: \nSilentTwo{} of the \nSilentTotal{} posts it
returned in the pilot named GPT Image 2 explicitly. Dropping this family reduces coverage but avoids admitting posts without
an explicit version signal.

Animated posts are kept: X serves GIFs as MP4, from which we extract frames at
four per second, drop consecutive near-duplicates and cap each animation at
twelve frames. The release contains \nGifFrame{} frames from \nGifAnimations{}
animations (one to \maxFramesPerGif{} each) alongside \nPhoto{} stills; frames
are marked in the manifest and correlated within an animation.

\section{Version attribution: caption expressions and three tiers}
\label{app:regex}

\begin{figure*}[t]
\centering
\begin{minipage}{\linewidth}\footnotesize\raggedright
\verb"2.5 signal:"\\
\verb"(image|images|gpt|chatgpt)[\s\-_]*2\.5|2\.5[\s\-_]*(flare|sunburst)|gptimage25|images25|"\\
\verb"image25|sunburst|flare|image2_5|2_5"\\[3pt]
\verb"2.0 signal:"\\
\verb"(gpt[\s\-_]*image|images?)[\s\-_]*2(\.0)?(?![\.\d_])|(?<![\d.])2\.0(?![\d])|"\\
\verb"gpt[\s\-_]*images?2(?![\d.\w])"
\end{minipage}
\caption{Case-insensitive version-gate expressions applied to post text for
caption-attributed records. Each signal is a single expression; the line break is
typographic. Host-attributed NightCafe records are gated on the platform's model
field instead.}
\label{fig:versiongate}
\end{figure*}

Caption-attributed records are scored by the following case-insensitive
regular expressions on post text. Host-attributed NightCafe records instead
require an allowed value in the platform's model field. Both signals are given in
Figure~\ref{fig:versiongate}. The 2.0 signal is written
in three alternatives: the prefixed form, whose negative lookahead keeps
``Images 2.5'' from firing as 2.0; a bare ``2.0'', which in a corpus already
naming 2.5 is the prior model and is what catches Chinese and Japanese
comparison posts, where no English prefix precedes the number; and the compact
\texttt{gptimage2}, guarded against ``GPTimage2.5''.

The bare and compact 2.0 alternatives exclude comparison posts that name both
versions. The stored caption records were re-gated with these expressions;
\texttt{regate.py} applies the rule by attribution route: caption records use
these expressions (and the stored classifier decision for official X), while
NightCafe records retain admission only when the recorded endpoint, field and
model value agree with an allowed host-model record. Host titles are not used
as a fallback version signal.

\subsection{Three attribution tiers}
\label{app:tiers}

The release identifies three tiers in \texttt{gate\_tier}: official-X caption
attribution plus the April creation-language classifier (\nOfficialTier{}
images); caption-regex-only attribution (\nCaptionTier{} images); and
host-recorded model attribution (\nHostTier{} images). The last tier also
contains \nHostVideoTier{} separately labelled videos, excluded from image
analyses.

NightCafe records are admitted when \texttt{gptImageModel} names
\texttt{gpt-image-2-5-sunburst} or \texttt{gpt-image-2-5-flare}; caption silence
is permitted in that tier. The host field is a platform assertion, not
independent verification that the endpoint executed. The collector excludes
records with its checked NSFW or underage moderation flags before curation.

Re-scoring the \nCaptionTier{} caption-only images with the official tier's
text classifier admits \nLooseTierWouldPass{} (\pctLooseTierWouldPass\%);
\nLooseTierNoCreation{} lack creation language it recognises. This is a
comparison of automated rules, not a validation: the classifier's phrase lists
are English- and Japanese-led, and missing language fields limit its use.
The three tiers must therefore remain selectable rather than interpreted as
interchangeable evidence of generator identity.

\section{Deduplication and cross-source overlap}
\label{app:dedup}

Runs are merged in the order pilot, official day one, gateway X, Xiaohongshu,
Weibo, Pixiv, note.com, Reddit, Instagram and NightCafe; the first occurrence wins. Exact
repeats are removed on the (post, media, frame) key (\nDupKey{});
near-duplicates are removed by a \hashBits-bit average hash (8$\times$8
grayscale) at Hamming distance $\le$\hashThresh{} against every image kept so
far (\nDupImage{}).

Replaying that pass over the run directories splits the
\nDupImage{} into \nDupSamePost{} images whose surviving match came from the
\emph{same} post and \nDupCrossPost{} from a different post, \nDupGifFrame{} of
them GIF frames.

Of the \nDupSamePost{} same-post drops, \nDupSamePostCrossRun{}
are the same post re-fetched through a second run (gateway rows carry no
\texttt{media\_key}, so only the hash stage catches them); the remaining
\nDupSamePostSameRun{} (\nDupSamePostSameRunStills{} stills,
\nDupSamePostSameRunGif{} GIF frames) are within-run same-post near-duplicates,
possible candidates for a before-and-after pair posted as two attachments, although only
\nDupSamePostSameRunStillsX{} of those stills are X rows, the rest being
carousel near-duplicates from the gallery-shaped sources.

The pass did not exempt
any of them. Treating every such X still as an edit-pair member gives an
unverified upper-bound scenario of \nDupSamePostSameRunStillsX{} losses (Appendix~\ref{app:limitations}).

The merge order also explains why the day-one run's \nOfficialStrictTweets{}
gate-passing posts become \nDayOnePostsSurviving{} in the merged manifest:
\nDayOnePostsInPilot{} were already carried by the pilot and lose the tie to it,
and \nDayOnePostsLostEntirely{} lost every image to hash collisions. The merged
manifest therefore records no cross-run overlap; the pre-dedup overlap of
Appendix~\ref{app:overlap} comes from the \texttt{dedup\_audit.py} script in the
collection code.

The pre-dedup overlap across \emph{sources} is measured separately:
\texttt{analysis/also\_seen\_in.py} recomputes which other sources hold a
near-duplicate of each released image, using the same \hashBits-bit average hash
at Hamming distance $\le$\alsoSeenThresh{} the merge itself uses, so it
reproduces the merge's similarity criterion, not semantic identity.

\nAlsoSeen{} of \nImages{}
released images (\pctAlsoSeen\%) were also seen on at least one other source:
\nSrcTwo{} on two, \nSrcThree{} on three, \nSrcFour{} on four and
\nSrcFive{} on five. That script
computes cross-source matches for the released images and stores them in a
separate analysis output; these matches are not a column in
\texttt{metadata.csv}.

\section{The screenshot filter: prompt, cost and labels by source}
\label{app:prompt}

Every gated image is labelled with \filterModel{} through the Responses API
(single-pass labels with default sampling, accumulated through the snapshot) at \texttt{detail=low} on a
768-pixel JPEG thumbnail, which cost \filterTokensIn{} input and
\filterTokensOut{} output tokens over \nFilterCalls{} calls
(\filterTokensPerImage{} input tokens per labelled row; the call count exceeds the
\nLabeled{} gated images because the re-gate discarded rows the filter had
already labelled).

At the model's list price of \$\priceFilterIn{} per million
input and \$\priceFilterOut{} per million output tokens, that is \$\costFilter{}
for the whole pass, or \$\costFilterPerK{} per thousand images. The fixed prompt asks for exactly one of six labels:

{\footnotesize\begin{list}{}{\leftmargin=0.6em\rightmargin=0pt}\item\ttfamily\raggedright\sloppy\setlength{\parskip}{0pt}
You are curating a dataset of images generated by ChatGPT Images 2.5. Classify this image into exactly one label:
standalone\_generated: a single image as the model would output it. No app chrome, no browser, no message bubbles, not a phone/desktop screenshot.
chat\_screenshot: a screenshot of ChatGPT / a chat app / a browser / a phone UI that contains a generated image inside it.
collage\_or\_grid: two or more separate images stitched together (before/after, 2x2 grid, candidates side by side).
photo\_of\_screen: a camera photo of a monitor or phone displaying an image.
promo\_graphic: a marketing/announcement graphic (platform logo, ``now available'', pricing tables, feature lists).
other: anything else.
Respond with JSON only: \{"label": ..., "confidence": "high|medium|low", "why": ...\}
\end{list}}

\subsection{Labels by source}
\label{app:screentable}

Table~\ref{tab:screen} gives the full breakdown summarised in
Section~\ref{sec:screen}. Instagram sits near the Xiaohongshu end (\nIgPromo{}
of \nSeenIg{} promotional). The collection design does not establish whether this reflects
platform composition, the hashtag query, or other selection effects. Because the
sources were queried with different instruments
(Appendix~\ref{app:instruments}), that contrast describes what our queries
returned in the collection window, not the platforms' populations. Within X the official
tier is also \pctPromoXOfficial\% promotional against \pctPromoXGateway\% for
the gateway rows, so part of every cross-source promotional contrast is gate
tier, not platform.

\begin{table*}[h]
  \centering\footnotesize
  \caption{Screenshot-filter labels by source over all gated images; only the
  standalone column gives automated admissions before the \nReviewedExcluded{}
  targeted review exclusions; the last column is that automated share. X is split by
  gate tier: official-API rows also passed the creation-language classifier,
  gateway and other caption-attributed rows passed the version regex alone;
  NightCafe uses the host model field. Videos are excluded. Reddit's \nSeenRd{} gated
  images are too few to read as a rate.}
  \label{tab:screen}
  \begin{tabular}{lrrrrrrrr}
    \toprule
    \textbf{Source} & \textbf{Gated} & \textbf{Standalone} & \textbf{Chat} & \textbf{Collage} & \textbf{Promo} & \textbf{Screen photo} & \textbf{Other} & \textbf{Standalone \%} \\
    \midrule
    X, official caption & 548 & 371 & 34 & 72 & 61 & 3 & 7 & 67.7 \\
X, caption regex & 2,091 & 1,130 & 195 & 246 & 447 & 31 & 42 & 54.0 \\
Xiaohongshu & 1,594 & 340 & 198 & 118 & 865 & 23 & 50 & 21.3 \\
Weibo & 165 & 40 & 26 & 10 & 80 & 4 & 5 & 24.2 \\
note.com & 262 & 138 & 48 & 25 & 43 & 2 & 6 & 52.7 \\
Pixiv & 72 & 66 & 0 & 2 & 3 & 0 & 1 & 91.7 \\
Instagram & 193 & 40 & 0 & 11 & 140 & 0 & 2 & 20.7 \\
Reddit & 18 & 11 & 1 & 2 & 3 & 1 & 0 & 61.1 \\
NightCafe & 1,401 & 1,358 & 0 & 20 & 19 & 0 & 4 & 96.9 \\
    \midrule
    Total       & \nLabeled{}  & \nLabStandalone{}  & \nLabChatShot{}  & \nLabCollage{}  & \nLabPromo{}  & \nLabScreenPhoto{}  & \nLabOther{}  & \pctStandaloneAll \\
    \bottomrule
  \end{tabular}

\end{table*}

\section{Historical audit design and detail}
\label{app:auditdetail}

The audits concern the 9 September curation, before the 10 September NightCafe
expansion. Retained per-item files record seed \auditSeed{}, identifiers,
sources and strata. They lack exact run timestamps and a complete immutable
sampling frame. We therefore report stored populations rather than reconstructing
them from the current manifest.

The screenshot frame contains \nAuditScreenPopulation{} images:
\nAuditScreenPopStandalone{} standalone,
\nAuditScreenPopChatShot{} chat screenshots, \nAuditScreenPopCollage{}
collages, \nAuditScreenPopPromo{} promotional graphics,
\nAuditScreenPopScreenPhoto{} screen photographs and
\nAuditScreenPopOther{} other. Respective sample sizes are
\nAuditScreenStandalone{}, \nAuditScreenChatShot{}, \nAuditScreenCollage{},
\nAuditScreenPromo{}, \nAuditScreenScreenPhoto{} and zero. Within each
sampled stratum, the recorded random-sampling rule gives inclusion probability
$n_h/N_h$.

The \nAuditScreen{} sampled images come from X (\nAuditScreenSrcTw{}),
Xiaohongshu (\nAuditScreenSrcXhs{}), Instagram (\nAuditScreenSrcIg{}),
note.com (\nAuditScreenSrcNote{}), Weibo (\nAuditScreenSrcWb{}) and Pixiv
(\nAuditScreenSrcPx{}). None comes from NightCafe. The taxonomy sample also
excludes NightCafe and contains \nAuditTaxTw{} X images among
\nAuditTax{} items. Its full historical frame size cannot be recovered from
the summary alone.

The hashes and reconstructed stratum metadata in
\texttt{audit-provenance.json} identify retained evidence, not the missing
historical frame. Neither audit establishes performance on the current cohort.

\subsection{Version gate, per language}
\label{app:gateaudit}

The gate audit sampled \nAuditGatePerLang{} posts in each language-by-decision
stratum. For English, Japanese, Chinese and other languages, respectively, the
stored admitted stratum sizes are \nAuditGateAdmPopEN{}, \nAuditGateAdmPopJA{},
\nAuditGateAdmPopZH{} and \nAuditGateAdmPopOTHER{}; rejected sizes are
\nAuditGateRejPopEN{}, \nAuditGateRejPopJA{}, \nAuditGateRejPopZH{} and
\nAuditGateRejPopOTHER{}. Sampling probabilities are therefore
\nAuditGatePerLang{} divided by these sizes.

Rejected posts exist only for official X. Within the admitted samples, the
numbers of official-X posts are \nGateAdmOfficialEN{}, \nGateAdmOfficialJA{},
\nGateAdmOfficialZH{} and \nGateAdmOfficialOTHER{}.

The stored summary's official-X
recall calculation reweights these realised subsets, although the original
random draw used combined-tier language-by-decision strata. We do not report
that quantity as a Horvitz--Thompson estimate. A design-consistent domain
analysis with uncertainty, and a sufficiently sized official-X sample, would
be needed before interpreting recall. Other tiers have no sampled rejected
population.

Of \nAuditGatePerLang{} rejected texts per language, the auditor calls
\nGateRejYesEN{} English, \nGateRejYesJA{} Japanese, \nGateRejYesZH{} Chinese
and \nGateRejYesOTHER{} other-language texts creation claims. These judgements
are about what text claims, not which generator produced the attached pixels.

\subsection{Screenshot filter, per class}
\label{app:filteraudit}

Among images predicted standalone in the historical sample, model disagreement
is \pctScreenContam\% (\pctScreenContamLo--\pctScreenContamHi).
Within other sampled classes, label agreement is
\pctScreenPrecCollage\% (\pctScreenPrecCollageLo--\pctScreenPrecCollageHi)
for \texttt{collage\_\allowbreak or\_\allowbreak grid},
\pctScreenPrecPromo\% (\pctScreenPrecPromoLo--\pctScreenPrecPromoHi) for
\texttt{promo\_\allowbreak graphic}, \pctScreenPrecChatShot\% for
\texttt{chat\_\allowbreak screenshot} and \pctScreenPrecScreenPhoto\% for the 38-image
\texttt{photo\_\allowbreak of\_\allowbreak screen} census. The intervals describe model-label
proportions in the sampled strata, not accuracy against ground truth.

Weighting by historical stratum sizes gives standalone recall only over the
five sampled strata. The unsampled 61-image \texttt{other} stratum could add
between zero and 61 auditor-standalone images; its contribution is unmeasured.
We do not interpret the stored whole-pool implied share, which assigns this
stratum no standalone contribution, as an identified population estimate.

\subsection{Subject taxonomy, per class}
\label{app:taxaudit}

Over \nAuditTax{} historically sampled images, \texttt{anime/illustrated}
agrees with the auditor on \pctTaxAnime\% of \nTaxClipAnime{} images;
\texttt{text-graphic} agrees on \pctTaxTextGraphic\% and
\texttt{product/UI mockup} on \pctTaxMockup\%. CLIP assigns
\texttt{text-graphic} \nTaxClipTextGraphic{} times versus the auditor's
\nTaxAuditorTextGraphic{}, and \texttt{photorealistic portrait}
\nTaxClipPortrait{} times versus \nTaxAuditorPortrait{}. These sample
contrasts motivate caution about class prevalence; they neither calibrate the
expanded release nor bound its true category shares.

\paragraph{Content-analysis settings.}
\label{app:contentsettings}
Subject classification uses CLIP ViT-L/14~\cite{radford2021clip} zero-shot over the
eight April category names with newly authored templates; the original April
templates were unavailable, so the backbone and category names do not establish an
identical classifier.

EasyOCR is routed by post-language metadata: Japanese labels
use Japanese and English, Chinese labels use simplified Chinese and English, and all
other or missing labels use English only, so the \nLangNull{} images without
language metadata all receive English-only OCR. Nonempty detections count at
confidence at least 0.3. Faces use InsightFace \texttt{buffalo\_l}.

Summaries cover
successfully analysed released non-video images; video decoding failures are
excluded rather than counted as zero detections. April's analyses ran on a set that
still contained screenshots and comparison grids, both text-bearing by construction,
so every April comparison is between a cleaned and an uncleaned set under differing
CLIP templates.

\subsection{Language field and creation claims}
\label{app:langaudit}

A separate historical text audit sampled \nAuditLangPer{} released posts per
language in English, Japanese and Chinese, using \auditorModel{} and seed
\auditSeed{}. The auditor agrees with the recorded language on
\pctLangAgreeAll\% of \nAuditLang{} sampled posts (English
\pctLangAgreeEN\%, Japanese \pctLangAgreeJA\%, Chinese \pctLangAgreeZH\%).
It reads \pctCreationAll\% (\pctCreationAllLo--\pctCreationAllHi) as the
author presenting an image they made: Japanese \pctCreationJA\%, English
\pctCreationEN\%, Chinese \pctCreationZH\%
(\pctCreationZHLo--\pctCreationZHHi). These are balanced-sample results;
the pooled rate does not estimate the current release's creation-claim share.
The audit does not cover records without a language field, including NightCafe,
and it does not validate the wording regexes as measures of user intent.

\section{Languages, timing and image geometry}
\label{app:lang}

Table~\ref{tab:lang} gives the language distribution of the gated pool against
the release. The \nXLang{} X rows carry the API's detected \texttt{lang}; the
\nPlatformDefaultLang{} rows (\pctPlatformDefaultLang\%) from Xiaohongshu and
Weibo (\nPlatformDefaultZH{}, \texttt{zh}) and Pixiv and note.com
(\nPlatformDefaultJA{}, \texttt{ja}) are assigned a platform default by
construction; the \nLangNull{} \noLangSourceList{} rows carry none, their
collectors returning no such field.

Comparison with April's \aprilPctJA\% Japanese share is affected by four
construction choices: differential filter retention, the addition of two
Japanese-first and two Chinese-first sources, platform-default labels, and
missing language fields on \pctLangNull\% of rows. Changes in early adoption
cannot be separated from these choices. The field also describes the post, not the image.

\begin{table*}[h]
  \centering\footnotesize
  \caption{Language distribution, gated pool against release. ``Platform
  default'' counts released rows whose language is a per-source constant rather
  than a detected field; the ``no field'' row includes NightCafe, Reddit and Instagram.
  Retention is the released share of the gated count.}
  \label{tab:lang}
  \begin{tabular}{lrrrrr}
    \toprule
    \textbf{Language} & \textbf{Gated} & \textbf{Released} & \textbf{of which X (detected)} & \textbf{of which platform default} & \textbf{Retention} \\
    \midrule
    English (en)  & \nGatedLangEN{} & \nLangEN{} & \nXLangEN{} & 0 & \pctRetainLangEN\% \\
    Japanese (ja) & \nGatedLangJA{} & \nLangJA{} & \nXLangJA{} & \nPlatformDefaultJA{} & \pctRetainLangJA\% \\
    Chinese (zh)  & \nGatedLangZH{} & \nLangZH{} & \nXLangZH{} & \nPlatformDefaultZH{} & \pctRetainLangZH\% \\
    Other (French \nLangFR{}, \nLangTailReal{} others, 1 undetermined) & \nGatedLangOther{} & \nLangOtherAll{} & \nLangOtherAll{} & 0 & \pctRetainLangOther\% \\
    No field (\noLangSourceList{}) & \nGatedLangNull{} & \nLangNull{} & 0 & 0 & \pctRetainLangNull\% \\
    \midrule
    Total & \nMergedImages{} & \nImages{} & \nXLang{} & \nPlatformDefaultLang{} & \pctRetainAll\% \\
    \bottomrule
  \end{tabular}
\end{table*}

\subsection{Timing and wording}
\label{app:timing}

\begin{figure*}[t]
  \centering
  \includegraphics[width=0.48\textwidth]{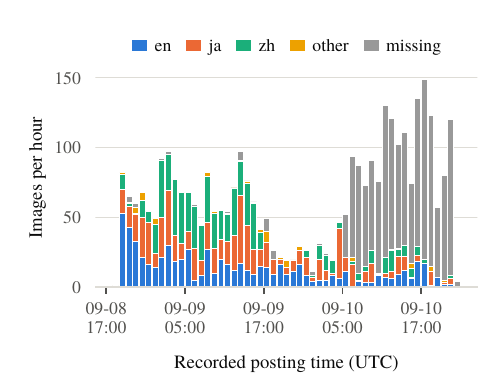}\hfill
  \includegraphics[width=0.48\textwidth]{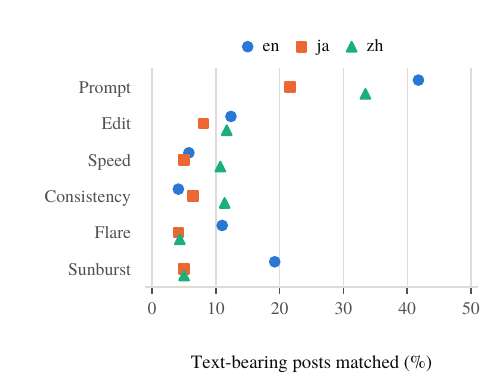}
  \caption{Left: released images by recorded posting hour in the \windowStart{}--\windowEnd{}
  window (\windowHours{} hours), stacked by recorded post language, with
  missing language shown separately. Differing source coverage prevents an
  adoption-rate interpretation. Right: share of
  released image posts with nonempty text whose text matches prompt-sharing,
  editing, speed, consistency
  and codename patterns, for English (\nPostsEN{} posts), Japanese
  (\nPostsJA{}) and Chinese (\nPostsZH{}); these are descriptive text-pattern
  matches, and the other recorded languages have too few posts to plot.}
  \label{fig:timing_wording}
\end{figure*}

Table~\ref{tab:source_timing} reports the earliest and latest posting timestamps
among each source's released images. These observed ranges do not establish
complete query coverage. For NightCafe, \texttt{created\_at} is normalised from
the host's \texttt{postedDate}; retrieval and generation times are different
quantities. The 10 September collection paginated the newest-items feed
retrospectively, so its retrieval date does not date every retrieved post.
Figure~\ref{fig:timing_wording} shows images by posting hour and caption
wording by language.

\begin{table*}[h]
  \centering\footnotesize
  \caption{Observed posting ranges among released images by source (UTC).
  Ranges describe retained records, not known complete collection windows or
  generation times.}
  \label{tab:source_timing}
  \begin{tabular}{lrll}
\toprule
Source & Images & First observed posting (UTC) & Last observed posting (UTC) \\
\midrule
X & 1,501 & 2026-09-08 18:58:45 & 2026-09-10 21:47:00 \\
Xiaohongshu & 325 & 2026-09-08 19:24:47 & 2026-09-10 12:28:43 \\
Weibo & 40 & 2026-09-09 00:26:00 & 2026-09-10 10:15:00 \\
note.com & 137 & 2026-09-09 02:15:11 & 2026-09-10 18:05:19 \\
Pixiv & 66 & 2026-09-08 21:11:51 & 2026-09-10 13:57:04 \\
Instagram & 40 & 2026-09-08 20:12:43 & 2026-09-10 17:53:47 \\
Reddit & 11 & 2026-09-09 02:36:14 & 2026-09-10 18:13:34 \\
NightCafe & 1,358 & 2026-09-10 05:25:55 & 2026-09-10 22:06:41 \\
\bottomrule
\end{tabular}

\end{table*}

\subsection{Image geometry as served}
\label{app:geometry}

Served dimensions are recorded for \nWithDims{} official-API rows
(\pctWithDims\%). The following statistics apply only to that subset.
Its images occupy \nDistinctSizes{} pixel sizes. Of these rows,
\nApiNative{} match an API-native size ($1024\times1024$,
$1024\times1536$ or $1536\times1024$), and \nChatExport{} match a
ChatGPT export size. Portrait images account for \pctPortrait\%, landscape
for \pctLandscape\% and square for \pctSquare\%.

Dimensions alone identify neither the original output nor later transformations.
Other sizes may result from editing, export or platform resampling, while native
sizes do not establish unchanged bytes. Generator-level frequency analysis
requires a verified delivery path and transformation history.

\section{Content credentials in detail}
\label{app:c2pa}

A pilot inspected \nCpaProbed{} Pixiv originals: \nCpaFound{} PNGs had
manifests whose signature chains were verified as OpenAI's, declaring trained
algorithmic media and a watermark assertion. The remaining three were JPEG
re-saves with no manifest detected. The correspondence between the two verified
pilot files and the current release has not been established, so the pilot does
not supply a verified-signature count for the current subset.

A byte scan of all \nPxCpaTotal{} released Pixiv files found C2PA markers in
\nPxCpaFound{} (\pctPxCpa\%). Of these, \nPxCpaNamesTwo{} matched the searched
\texttt{gpt-image} version \cpaGeneratorVersion{} string, an OpenAI string and
a watermark string. One marker-positive file lacked that combination and remains
uncharacterised. Byte-marker and string matches do not verify a manifest or its
signature.

The separate API ledger covers \nApiCpaTotal{} files from the three
recorded endpoints on \apiCpaDate{}; the inspected generator-version field is
\cpaGeneratorVersion{} throughout that ledger. This establishes a limitation
of that field in those files, not all provenance assertions or all endpoints.

Because the inspected field does not distinguish versions, it cannot confirm
that the caption-attributed Pixiv images were generated by 2.5. Their version
attribution remains caption-based. Manifest presence is not an admission signal
or a released CSV field. The expanded NightCafe cohort has not received the same
credential inspection, so we do not infer whether it preserves or strips C2PA.

\section{The source landscape, platform by platform}
\label{app:landscape}

\paragraph{Social and creator platforms.} X, Reddit, Weibo, Xiaohongshu,
Instagram and Threads expose caption-based evidence. Our source observations
come from probes on 9--10 September recorded in our collection notes, not an
exhaustive survey. Pixiv contributes \nKeptPx{} released images from
\nKeptPostsPx{} works; all \nPxAiDeclared{} carry its structured AI declaration.
The declaration is still supplied by the uploader and does not name a generator
version. Pixiv originals offer a byte-preserving delivery route in the inspected
files, whereas note.com and social CDN paths can re-encode images. We do not
extrapolate a tested delivery path to every asset on a platform. Credential
inspection coverage is given in Appendix~\ref{app:c2pa}.

\paragraph{Model-tagged galleries.} A host-recorded model field provides an
attribution route distinct from caption claims. NightCafe's public creations
endpoint exposes \texttt{gptImageModel}; the 9 September probe observed no 2.5
items among \nNightCafeSampled{} newest creations. The 10 September crawl
collected \nNcCreations{} creations bearing allowed 2.5 values. These snapshots
show a change in observed feed contents, not a precisely measured adoption time.

The frozen snapshot includes \nKeptNightcafe{} NightCafe images and
\nHostVideoTier{} separate videos. A caption-recall study could compare text
against these host labels within this gallery, but would still depend on their
validity and would not estimate recall on other platforms.

On 10 September, SeaArt's \nSeaArtCards{} model cards reported
\nSeaArtTasksSunburst{} and \nSeaArtTasksFlare{} generation tasks but
\nSeaArtWorksTotal{} published works. The imagine.art probe returned no items
for the 2.5 style identifiers \nImagineStyleFlare{} and
\nImagineStyleSunburst{}, nor among \nImagineNewest{} newest posts. Its
\texttt{style\_id} filter returned expected matches on control identifiers;
a separate \texttt{model\_id} parameter did not change the results and cannot
be used as attribution evidence. These are query-specific observations of no
items, not evidence of no platform adoption.

Mage.space's public model counts
rose from \nMageTotal{} on \probeDateOne{} to \nMageTotalTwo{} on
\probeDateTwo{} (\nMageSunburstTwo{} Sunburst, \nMageFlareTwo{} Flare), while
its per-image feed remained behind login and was not collected.

\paragraph{API resellers and aggregators.} The same survey found third-party
access routes including fal.ai, Kie.ai, OpenArt, Mage.space and Krea. Availability
of generation does not establish a public, attributable gallery. OpenArt's
observed model-filter results did not consistently carry a corresponding
per-item model field, so we did not admit them on that basis. Intermediary
re-encoding is a separate provenance concern that must be checked against the
actual returned bytes.

\paragraph{Collection implications.} Collect accessible social and creator
sources in parallel, recording their distinct gates and delivery paths. Poll
model-tagged galleries during the launch week: NightCafe demonstrates that a
single empty probe can miss a substantial later cohort. Preserve dates and query
parameters for negative observations, and keep host attribution separate from
caption attribution. Choice of an official API or gateway should depend on the
observed yield, cost assumptions and reproducibility of each route rather than
an assumed universal ranking of their coverage.

\section{What the release directory contains}
\label{app:release}

The prepared local release directory contains the image cohort and separate
video assets, a
flat \texttt{metadata.csv} with \nCsvColumns{} columns, and a datasheet. Each
row carries a content-hash identifier, post identifier and URL (the legacy names
\texttt{tweet\_id} and \texttt{tweet\_url} apply across sources), recorded posting time,
language, media type, frame index, served dimensions, confirmation route,
version signal, screenshot label, query family, \texttt{source\_platform},
\texttt{pixiv\_ai\_type}, like count and post text (the host's public creation
title for host-attributed rows, not a verified generation prompt).

The selectable
\texttt{gate\_tier} separates official-X caption plus classifier,
caption-regex-only, and host-recorded model attribution. The \texttt{source\_run} field records the surviving collection run. Host records
additionally carry \texttt{model\_id}, \texttt{attribution\_endpoint} and
\texttt{attribution\_field}. No author profile fields are exported.
The datasheet states the frozen window, image/video counts and three admission
rules. The release is hosted at \mbox{\url{https://scam.ai/research}}.
Analyses of images exclude the video rows using \texttt{media\_type}.

C2PA signature status, per-item human adjudication and full reconstruction of the
historical audit sampling frames are not provided. Cross-source hash matches
remain analysis outputs rather than an attribution guarantee. These limitations
should be retained when selecting a cohort or interpreting a released label.

\section{Aggregate subject comparison}
\label{app:taxonomy}

\begin{figure}[t]
  \centering
  \figph{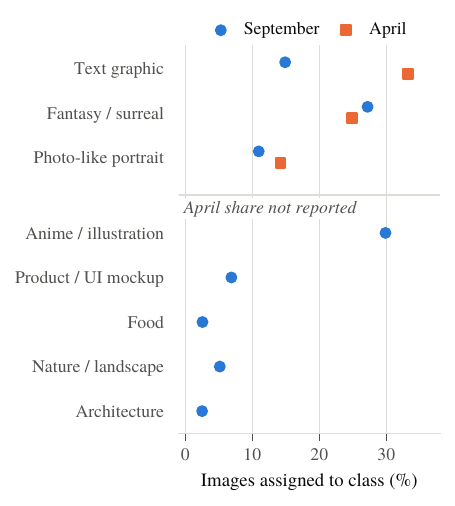}{1.0}
  \caption{Subject-matter distribution of the release under CLIP ViT-L/14
  zero-shot classification into the eight April categories using newly authored
  templates. April shares are
  shown for the three classes that paper reported, computed on an uncleaned set
  that still contained screenshots and grids. Read the shares as CLIP's: an
  earlier-cohort audit, excluding NightCafe, found $\kappa=\taxKappa{}$ and
  substantial disagreements on
  \texttt{text-graphic} and \texttt{photorealistic portrait}
  (Section~\ref{sec:audit}).}
  \label{fig:taxonomy}
\end{figure}

Figure~\ref{fig:taxonomy} compares CLIP-assigned subject shares with the
three categories reported in April. The differing templates and image-form
filters limit this descriptive comparison.

\section{Detector Evaluation Details}
\label{app:detectors}

\paragraph{Selection and controls.}
The experiment uses the public Community Forensics 384-pixel checkpoint,
B-Free's \texttt{BFREE\_\allowbreak dino2reg4/\allowbreak model\_\allowbreak epoch\_\allowbreak best.pth}, and Effort's
\texttt{genimage\_\allowbreak checkpoint.pth}, plus PGC's
\texttt{PGC\_\allowbreak train\_\allowbreak sdv1\_4\_\allowbreak ckpt.pth}, PROBE's
\texttt{ResNet50\_\allowbreak best\_\allowbreak model\_\allowbreak step\_\allowbreak 19343.pth}, and DoU's released \texttt{best.pth}.
Checkpoints are selected before inspecting collection scores.
Photograph pools derive from
\texttt{MLap/\allowbreak GeoDE}, \texttt{tanganke/\allowbreak sun397},
\texttt{HuggingFaceM4/\allowbreak FairFace}, \texttt{nateraw/\allowbreak pascal-voc-2012}, and
\texttt{ethz/\allowbreak food101}; artwork controls derive from \texttt{huggan/\allowbreak wikiart}.
Selection orders candidate paths by SHA256 of the fixed prefix
\texttt{paper-w-controls-v1:} followed by the path, retains the first 400
byte-distinct images per source, and alternates calibration/test assignment.
Manifests retain source, split, image identity, and byte hashes.

\paragraph{Execution and checks.}
All inference is FP32 with TF32 disabled. Community Forensics uses its official
image processor at 384 pixels; B-Free preserves native patch processing;
Effort uses RGB conversion, OpenCV linear resizing to $224\times224$, and CLIP
normalization. B-Free's batched adapter and Effort's folded frozen SVD weights
matched upstream CPU forward outputs exactly on the recorded validation cases;
Effort preprocessing also matched upstream on 112 pilot images. Models run
sequentially on one RTX 3090, using batches of 64, 8, and 32 and respectively
six, four, and four preprocessing workers with pinned memory and prefetching.

PGC preserves the official native-pixel padding and center crop at 224 pixels,
appends three quantization-residual channels, and normalizes RGB and residual
channels separately. It uses batches of 32 with four preprocessing workers.
Its strict load consumed all 563 checkpoint keys, and its transform and forward
path matched the upstream evaluator exactly on four recorded input images.

PROBE-ResNet50 strictly loads its 320 checkpoint keys and retains native
224-pixel patch extraction, averaging patch logits before sigmoid. Small inputs
use the official black-padding policy. Image batches of eight feed GPU
microbatches of 32 patches.

DoU retains its official CLIP-based transform
and stochastic forward, using batches of 32. Both use four preprocessing workers.
Their adapters were checked against upstream transforms and forwards.

The September-plus-controls runs take approximately 67, 644, 78, 78, 154, and 75 seconds. All 35,268 primary scores are
finite, with exact expected image-ID coverage. Recorded checkpoint, code, and
manifest hashes accompany the score files.

\paragraph{Calibration and uncertainty.}
For sorted calibration logits $s_{(1)}\le\cdots\le s_{(n)}$, the threshold is
set to the next representable double above $s_{(n-\lfloor0.05n\rfloor)}$;
scores at or above the threshold are flagged. This rule conservatively handles
ties and flags exactly 50 of 1,000 calibration photographs for each detector.
Table~\ref{tab:detectors} and the source-level rates use 1,000 post-clustered bootstrap draws;
the cross-cohort comparison uses \xvDraws{} draws per cohort.

Held-out photograph
FPR Wilson intervals are 4.6--7.5\%, 3.1--5.6\%, 2.5--4.8\%, 3.9--6.6\%, 4.1--6.9\%, and 3.3--5.9\%; artwork
intervals are 35.4--48.9\%, 57.1--70.3\%, 5.4--13.2\%, 93.0--98.3\%, 0.5--4.3\%, and 7.4--16.1\%, in table order.
All intervals condition on the fitted calibration thresholds.

\paragraph{Overlap sensitivity.}
No byte-identical overlap was found between collection and control inputs, and
no decoded-pixel identity was found across the control calibration/test splits.
A dHash distance screen ($\le3$) identified three cross-split FairFace candidate
pairs, not established duplicates. Removing all six candidate images and
recalibrating on 997 photographs yields held-out photograph FPRs of 6.02\%,
4.21\%, 3.51\%, 5.02\%, 5.12\%, and 4.31\% on 997 images, respectively. Residual perceptual overlap
and detector-training overlap are not excluded by these checks.

\paragraph{Exploratory content restriction.}
Restricting the collection to 736 images assigned the existing CLIP categories
portrait, food, nature/landscape, or architecture gives AUROCs of 0.841, 0.535,
0.929, 0.846, 0.538, and 0.695 against the 1,000 held-out reference photographs. These are
classifier-defined content proxies, not human-validated photographic images or
semantically matched controls. The comparison remains secondary to the
source-resolved flag rates and does not independently validate image origin.

\paragraph{DoU stochastic evaluation.}
DoU samples latent Gaussian noise in its published evaluation forward pass.
We preserve this behavior, using seed 42 for the primary run and recording
batch size 32. A second full pass with seed 43 checks sensitivity without
changing the primary result. With each pass calibrated on the same reference
photograph split, collection flag rates are 17.4\% and 17.9\%, and held-out
photograph FPRs are 4.4\% and 6.9\%.

Holding the primary threshold fixed instead
gives a seed-43 flag rate of 17.8\% and photograph FPR of 6.6\%; 235 of 3,478
collection images change verdict.

These two realizations do not estimate the
full distribution over seeds. Post-clustered bootstrap intervals omit this additional
stochastic variation. Partial resume is rejected rather than silently reseeding
and skipping previously scored images.

\begin{table*}[htbp]
\centering\small
\caption{Collection flag rates (\%) by source, with half-widths of 95\%
post-clustered bootstrap intervals. CF is Community Forensics and PROBE is
PROBE-ResNet50. Each detector uses the same
photograph-calibrated threshold across all sources. Intervals
resample posts and condition on the fitted threshold. Small sources are thin:
Reddit contributes 11 images from 4 posts. Differences describe the collected sample, not the causal effect of publishing
on a platform. DoU uses seed 42; intervals omit variability from its latent-noise
sampling.}
\label{tab:detector_sources}
\begin{tabular}{lrrrrrrrr}
\toprule
Source & Images & Posts & CF & B-Free & Effort & PGC & PROBE & DoU \\
\midrule
X & 1,501 & 861 & 44.3\,\pmhw{3} & 19.3\,\pmhw{3} & 28.8\,\pmhw{3} & 43.2\,\pmhw{3} & 3.8\,\pmhw{1} & 11.1\,\pmhw{2} \\
NightCafe & 1,358 & 1,358 & 64.1\,\pmhw{3} & 8.2\,\pmhw{1} & 52.6\,\pmhw{3} & 56.0\,\pmhw{3} & 2.4\,\pmhw{1} & 3.5\,\pmhw{1} \\
Xiaohongshu & 325 & 133 & 50.8\,\pmhw{6} & 20.3\,\pmhw{5} & 61.5\,\pmhw{8} & 96.9\,\pmhw{2} & 3.1\,\pmhw{3} & 53.8\,\pmhw{6} \\
note.com & 137 & 29 & 70.1\,\pmhw{13} & 6.6\,\pmhw{5} & 78.1\,\pmhw{11} & 92.7\,\pmhw{6} & 1.5\,\pmhw{2} & 93.4\,\pmhw{5} \\
Pixiv & 66 & 18 & 81.8\,\pmhw{16} & 45.5\,\pmhw{31} & 37.9\,\pmhw{25} & 84.8\,\pmhw{12} & 31.8\,\pmhw{31} & 60.6\,\pmhw{36} \\
Instagram & 40 & 18 & 35.0\,\pmhw{19} & 10.0\,\pmhw{15} & 47.5\,\pmhw{22} & 47.5\,\pmhw{24} & 10.0\,\pmhw{13} & 17.5\,\pmhw{14} \\
Weibo & 40 & 19 & 70.0\,\pmhw{18} & 22.5\,\pmhw{16} & 45.0\,\pmhw{25} & 62.5\,\pmhw{22} & 5.0\,\pmhw{8} & 77.5\,\pmhw{22} \\
Reddit & 11 & 4 & 90.9\,\pmhw{15} & 0.0\,\pmhw{0} & 45.5\,\pmhw{50} & 81.8\,\pmhw{30} & 0.0\,\pmhw{0} & 81.8\,\pmhw{30} \\
\midrule
All sources & 3,478 & 2,440 & 54.7\,\pmhw{2} & 15.0\,\pmhw{2} & 43.7\,\pmhw{2} & 56.4\,\pmhw{2} & 3.7\,\pmhw{1} & 17.4\,\pmhw{2} \\
\bottomrule
\end{tabular}

\end{table*}

\subsection{Harness reproduction on GenImage}
\label{app:reproduction}

\paragraph{Purpose of the benchmark check.}
Forward-pass parity and control FPRs test different parts of the implementation,
but neither validates the full scoring pipeline. Each adapter matches the
official forward pass bit for bit on identical tensors. Both sides, however,
receive already-preprocessed inputs, so this check cannot detect a shared
resizing error.

Held-out photograph FPRs of \pctFprLow--\pctFprHigh\% show that the scores
can be calibrated on real photographs. They include no generated image with
a known label. Together, these checks can still miss errors such as inverted
score direction or an omitted centre crop, which can yield plausible flag
rates on an unlabelled collection.

\paragraph{Sample and comparability.}
GenImage~\cite{genimage2023} is a common benchmark for these detectors.
Community Forensics reports results on it, Effort's checkpoint targets it,
and PROBE and PGC train on its SD~v1.4 split. We evaluate a frozen validation
sample with \nGenImagePerClass{} real and \nGenImagePerClass{} generated
images from each of \nGenImageGenerators{} available generators
(\nGenImageImages{} images). It is intended to detect large implementation
errors while accepting the sampling uncertainty of a small subset. No fitting
or tuning uses this sample.

\paragraph{Frozen sample and metric.}
Within each generator and class, files are ranked by SHA256 of
\texttt{paper-w-genimage-v1:} followed by generator, class and filename.
We retain the first \nGenImagePerClass{}. Selection reads filenames only
and finishes before inference. Our copy contains \nGenImageGenerators{}
of the eight published subsets; SD~v1.4 is absent. The training-domain column
for PROBE and PGC is therefore not included.

Accuracy uses each detector's own $0.5$ boundary. Equal real and generated
counts make accuracy identical to balanced accuracy within each generator.
We recompute published means over the same available generators. Community
Forensics is an exception: its published aggregate spans all eight, including
the missing SD~v1.4 subset, so that comparison is not exactly matched.

\paragraph{Published references.}
Community Forensics, PROBE, PGC and Effort references come from their authors:
\cite{cf2025} Table~2, \cite{cao2026probe} Table~2,
\cite{zhou2026pgc} Table~2 and \cite{effort2025} Table~9, respectively.
B-Free reports only averages on a recompressed GenImage variant; its reference
comes from the PGC study, \cite{zhou2026pgc} Table~2. DoU reports GenImage
results only graphically.

Reference provenance matters. Third-party Effort figures are more than ten
points below its own reported results. Using them would imply a roughly
fifteen-point discrepancy where the comparison with the authors' result is
much closer.

\begin{table*}[htbp]
\centering\footnotesize
\caption{Per-generator accuracy (\%) on the frozen GenImage sample and
published reference values. Agreement in per-generator patterns provides
additional evidence beyond the mean; it does not prove checkpoint identity.}
\label{tab:reproduction_detail}
\begin{tabular}{llrrrrrrrr}
\toprule
Detector & & ADM & BigGAN & Midjourney & SD1.5 & VQDM & Wukong & Glide & Mean \\
\midrule
Community Forensics & ours & 93.0 & 87.5 & 85.3 & 100.0 & 100.0 & 99.7 & 99.7 & 95.0 \\
 & published & -- & -- & -- & -- & -- & -- & -- & -- \\
\addlinespace[1pt]
B-Free & ours & 78.0 & 67.7 & 95.8 & 99.2 & 88.3 & 97.5 & 84.2 & 87.2 \\
 & published & 78.1 & 68.6 & 94.9 & 98.9 & 88.3 & 98.8 & 83.8 & 87.3 \\
\addlinespace[1pt]
Effort & ours & 84.8 & 92.7 & 84.2 & 98.3 & 93.7 & 98.0 & 93.2 & 92.1 \\
 & published & 78.7 & 77.6 & 82.4 & 99.8 & 91.7 & 97.4 & 93.3 & 88.7 \\
\addlinespace[1pt]
PGC (2026) & ours & 100.0 & 98.2 & 100.0 & 100.0 & 100.0 & 100.0 & 100.0 & 99.7 \\
 & published & 100.0 & 86.7 & 100.0 & 100.0 & 100.0 & 100.0 & 100.0 & 98.1 \\
\addlinespace[1pt]
PROBE-ResNet50 & ours & 57.0 & 49.8 & 92.0 & 99.3 & 62.3 & 99.7 & 60.8 & 74.4 \\
 & published & 60.0 & 49.4 & 90.7 & 99.2 & 66.9 & 99.4 & 63.0 & 75.5 \\
\addlinespace[1pt]
DoU & ours & 83.3 & 75.3 & 72.3 & 100.0 & 98.2 & 99.5 & 92.5 & 88.7 \\
 & published & -- & -- & -- & -- & -- & -- & -- & -- \\
\addlinespace[1pt]
\bottomrule
\end{tabular}

\end{table*}

\paragraph{Residual differences.}
Each generator contributes \gimPerGenN{} images, compared with twelve
thousand in the full validation split. Our estimates are therefore noisier and
use a different composition. Differences above one point run both ways: ours
is higher in six cells, mostly Effort, and lower in five, mostly PROBE-ResNet50.
The generated report
records per-generator values, differences and standard errors.

\paragraph{Community Forensics evaluation set.}
We also evaluate Community Forensics on a separately released, labelled
held-out set~\cite{cf2025}. A frozen balanced sample contains
\nCfEvalImages{} images (\nCfEvalPerLabel{} per class) across
\nCfEvalGenerators{} generators. Files are copied byte for byte from shards
containing \nCfEvalPopulation{} released images.

The harness reaches pooled average precision (AP) \cfMeasuredAp{} and
accuracy \cfMeasuredAccPct{}\%, compared with published mean AP
\cfPublishedMap{} and accuracy \cfPublishedAcc{}. The release does not
fully specify the published AP aggregation. Averaging per-generator AP on our
sample gives \cfMeasuredMap{}, rather than pooled AP \cfMeasuredAp{}.
Each generator is then compared with the entire real pool, producing generated
prevalence of roughly \pctCfGenPrevalence\%. Because AP depends on
prevalence, this is not directly comparable with the balanced published evaluation.

Pooled AP on the balanced sample is the closest comparison we can construct.
It falls about half an AP point below the published value, which lies outside
our bootstrap interval [\cfApCiLow{},~\cfApCiHigh{}]. Accuracy likewise
falls short, and the residual gap is unresolved. Averaging within released
shards gives \cfShardAp{} and does not close the gap.

The harness calls the authors' image processor in test mode, so preprocessing
is not independently reimplemented. Our sampling and the published pairing of
real and generated images remain possible sources of difference. On this same
set, the other five detectors reach AUROCs of at least \cfWorstAuroc{}.
Those out-of-distribution results lack comparable published references. We
report them only as evidence against inverted or chance-level scoring.

\paragraph{Metric computation.}
AP is pooled unless otherwise stated. Community Forensics evaluation intervals
use 2,000 image-level bootstrap draws with seed 20260912. This set has no
post-level grouping analogous to the online collection. Comparisons with
balanced published sets use balanced samples because AP depends on prevalence.

The AP and AUROC implementations are checked against \texttt{scikit-learn}
in our test suite, using continuous, integer-tied and
rounded scores. Ties matter for saturated detectors: cumulative-sum AP that
arbitrarily separates equal scores can disagree with the standard metric.
Such an error would otherwise resemble disagreement with a published result.

\paragraph{Scope.}
A released checkpoint can differ from the one used for a paper's table.
These checks cannot separate that possibility from a harness fault, so agreement
supports implementation correctness without proving it. They also do not
validate generator attribution or platform-specific preprocessing in the
collected images.

\subsection{Historical-cohort comparison}
\label{app:cross_version}

\begin{table*}[t]
\centering\small
\caption{Sensitivity to calibration sampling. Differences are September X
minus April X, in percentage points. Fixed-threshold intervals reproduce the
primary analysis; the other intervals additionally resample calibration
photographs within source and refit the threshold. Both analyses resample whole
posts and use 50,000 draws. The 99.17\% intervals adjust for six detectors;
DoU's latent-sampling variability is excluded.}
\label{tab:calibration_sensitivity}
\begin{tabular}{lrrrr}
\toprule
Detector & Difference & Fixed threshold & \multicolumn{2}{c}{Resampled calibration} \\
 & (pp) & 95\% interval & 95\% interval & 99.17\% interval \\
\midrule
Community Forensics & $-$0.5 & [$-$4.1, +3.2] & [$-$4.0, +3.4] & [$-$5.3, +4.6] \\
B-Free & +1.3 & [$-$1.7, +4.3] & [$-$1.8, +4.2] & [$-$2.9, +5.3] \\
Effort & +6.1 & [+2.7, +9.6] & [+2.8, +10.0] & [+1.7, +11.3] \\
PGC (2026) & +0.8 & [$-$3.1, +4.7] & [$-$2.9, +5.4] & [$-$4.4, +6.8] \\
PROBE-ResNet50 & +0.9 & [$-$0.3, +2.2] & [$-$0.3, +2.3] & [$-$0.7, +2.8] \\
DoU & +3.6 & [+1.0, +6.2] & [+1.1, +6.3] & [+0.3, +7.4] \\
\bottomrule
\end{tabular}

\end{table*}

\paragraph{Frozen April input.}
We retrieve \href{https://scam.ai/research}{the published GPT-Image-2 dataset}
at revision \path{a8e3f74cd02edcab94d8c61ae0439a9dc5d78468}, preserving all
\aprilImages{} metadata rows. The archive places 219 images in
\path{images_overflow/} although metadata paths point to \path{images/}.
Unique filenames resolve them without substituting files. The manifest retains
release order, post IDs, admission routes, languages, classifications and
both byte and decoded-RGB hashes.

Neither the sample nor the complete \aprilImages{}-image release overlaps
September images or reference controls at either hash level. We do not apply
September's screenshot filter retrospectively. Original classifications are
\nAprilConfirmed{} \texttt{confirmed} and 5,258 \texttt{uncertain} images;
recorded routes include \nAprilBadge{} generic AI-badge admissions and
\nAprilNameOnly{} name-only admissions. These labels do not authenticate
generator identity.

Posts span 21--28 April 2026. Recomputed image-row language counts are 4,117
English, 3,355 Japanese and 1,959 Chinese: \aprilPctEN\%, \aprilPctJA\%
and \aprilPctZH\% of \aprilImages{} rows. These reproduce the April
shares cited in this appendix. We do not pool them with September language
fields, whose sources, defaults and missingness differ.

\paragraph{Sample selection.}
All six detectors use the same \nAprilSample{} images from
\nAprilSamplePosts{} posts. We rank the release by SHA256 of
\texttt{paper-w-april-pilot100-v1:} followed by image ID and retain the first
\nAprilSample{} distinct records. The prefix is inherited from the initial
pilot and kept unchanged for reproducibility. Inference follows original
release order.

Selection precedes scoring and uses no quotas or score-based exclusions.
It samples image records without replacement but does not guarantee balanced
content or languages. Parent and sample manifest hashes and selected IDs are
recorded. The report checks score and manifest hashes, checkpoint identity,
preprocessing, score definitions, package versions and batch configuration
against September. Thresholds come from the saved September calibration report
and are never refitted on April.

\paragraph{Intervals and multiplicity.}
Each bootstrap draw samples source--post pairs with replacement within a
cohort and retains all images of each selected post. The rate is the resampled
flag count divided by the resampled image count. Cohorts are resampled
independently, and difference intervals use percentiles of draw-wise differences.
The report uses seed 20260912, deterministic detector-specific streams and
\xvDraws{} draws to estimate the family-wise tail bounds.

The six contrasts are exploratory. We report pointwise intervals and, for the
two excluding zero, Bonferroni-adjusted intervals for the family of
\xvNDetectors{}. Shared creators and residual image similarity can produce
dependence beyond the post clusters.

\paragraph{Attribution sensitivity.}
The metadata-only view restricts April to sampled records labelled
\texttt{confirmed}, a label held by \nAprilConfirmed{} of
\aprilImages{} parent images. September is restricted to \nOfficialTier{}
X images in \texttt{official\_caption\_classifier}. These admission rules
are neither matched nor independently verified. We do not compute difference
intervals for this view. Per-cohort intervals can rely on as few as seven
PROBE-ResNet50 flags, for which percentile-bootstrap coverage is unreliable.

\paragraph{Duplicate and seed sensitivity.}
The \nAprilSample{} records contain 17 byte-duplicate groups with 19 extra
images; the parent release has 122 groups and 145 extra images. The
pre-specified first-occurrence sensitivity retains 2,981 records. It changes
every flag rate by less than 0.1~pp: Community Forensics 44.80 to 44.78,
Effort 22.70 to 22.61, and DoU 7.57 to 7.48. No conclusion changes.

DoU uses primary seed 42 on April. The September-only second-seed check does
not quantify variability in the cross-cohort difference. Neither a favourable
seed nor an attribution subset replaces the primary comparison.

\paragraph{Resampling the calibration photographs.}
We add a post hoc sensitivity analysis after the primary comparison. Each of
50,000 draws resamples 200 calibration photographs within each of the five
source pools, preserving the fixed source allocation, and refits the same
conservative 5\% threshold. It independently resamples April and September
posts, retaining every image in each sampled post. Both cohorts use the same
refitted threshold in that draw. Seed 20260913 and separate detector-specific
streams make the calculation reproducible. The point estimates remain those
at the original fitted thresholds.

Table~\ref{tab:calibration_sensitivity} reports all six detectors, with
pointwise 95\% and Bonferroni 99.17\% intervals. Effort and DoU remain
positive under the family-wise adjustment; all other intervals include zero.
This analysis addresses calibration sampling conditional on the five chosen
photograph pools. It does not account for alternative control sources,
DoU's stochastic forward pass, selection bias or attribution error.

\subsection{Further limitations of the collection}
\label{app:limitations}

\paragraph{Attribution remains unverified.} Captions and host fields can misname
the generator; the inspected C2PA generator-version field does not resolve
that uncertainty. X's rendered ``Made with AI'' badge was unavailable in the
collection environment. Pixiv's AI declaration, present for
\nPxAiDeclared{} images, identifies AI use rather than generator version.

\paragraph{Historical audits do not validate the current cohort.}
The three automated stages were assessed by another model without human
adjudication. Correlated errors can remain. Screenshot and taxonomy audits
contain no NightCafe examples; the caption audit cannot assess host attribution.
Language quotas and unequal class sampling also limit pooled proportions.
Section~\ref{sec:audit} specifies the populations and estimands. A human audit
covering the frozen cohort and all attribution tiers remains future work.

\paragraph{Selection and source comparability.} Caption admission omits
version-silent posts; host-attributed galleries use a different selection rule.
Queries, time filters, budgets and image-form retention differ by route, so
source yields do not measure relative platform activity. Four platforms have
default language labels and three have none. Source and attribution fields
permit restricted analyses but do not remove these confounders.

\paragraph{Near-duplicate removal may discard distinct edits.} Average-hash
similarity does not establish semantic identity. Treating all
\nDupSamePostSameRunStillsX{} within-run, same-post X still-image drops as
edit-pair members gives an unverified upper-bound scenario for that subset
(Appendix~\ref{app:dedup}). Actual edit losses and gallery drops remain
unreviewed. Caption self-report also does not reliably distinguish generation
from editing.

\paragraph{Snapshot coverage.} Neither X route, nor their union, is a census.
The gateway's prices, endpoints and retrieval behaviour can change. Reddit and
Instagram were collected later, and some qualifying posts lack downloadable
media. NightCafe was added later still: its host posting timestamps span only
the final \ncWindowHours{} hours (\ncWindowStart{} to \ncWindowEnd{}).
It supplies \nLateNightCafe{} of \nLateImages{} images posted in that span.
These are publication times, not retrieval or verified generation times.
The window ends at \windowEnd{}, about \windowHours{} hours
after the anchor, and covers only part of launch week.

\paragraph{Content measurement.} Summaries exclude \nVideo{} videos and failed
image analyses. April comparisons mix source populations, image-form filters
and newly authored CLIP templates, despite a shared backbone and category names.
OCR readers are selected from post-language metadata; missing or other languages
receive English-only OCR. Post language can differ from image script, so
unsupported scripts may be missed. The detected share is not a validated
estimate of text prevalence.

\end{document}